\documentclass[10pt,twocolumn,letterpaper]{article}

\usepackage{algorithm}
\usepackage{algorithmic}
\usepackage{multirow}
\usepackage[accsupp]{axessibility}
\usepackage{cvpr}              
\definecolor{cvprblue}{rgb}{0.21,0.49,0.74}
\usepackage[pagebackref,breaklinks,colorlinks,allcolors=cvprblue]{hyperref}

\def\paperID{36656} 
\def\confName{CVPR}
\def\confYear{2026}

\title{StreamingTOM: Streaming Token Compression for Efficient Video Understanding}

\author{
Xueyi Chen\textsuperscript{1,2},
Keda Tao\textsuperscript{1,3},
Kele Shao\textsuperscript{1,4,3},
Huan Wang\textsuperscript{1,*} \\
\textsuperscript{1}Westlake University,
\textsuperscript{2}The Chinese University of Hong Kong,
\textsuperscript{3}Zhejiang University,
\textsuperscript{4}SII\\
\textsuperscript{*}Corresponding author: \texttt{wanghuan@westlake.edu.cn}\\
\url{https://yige24.github.io/StreamingTOM}
}

\begin{document}
\twocolumn[{
\renewcommand\twocolumn[1][]{#1}
\maketitle
\begin{center}
\vspace{-9mm}
\begin{tabular}{@{}c@{}c@{}}
\includegraphics[width=0.637\linewidth]{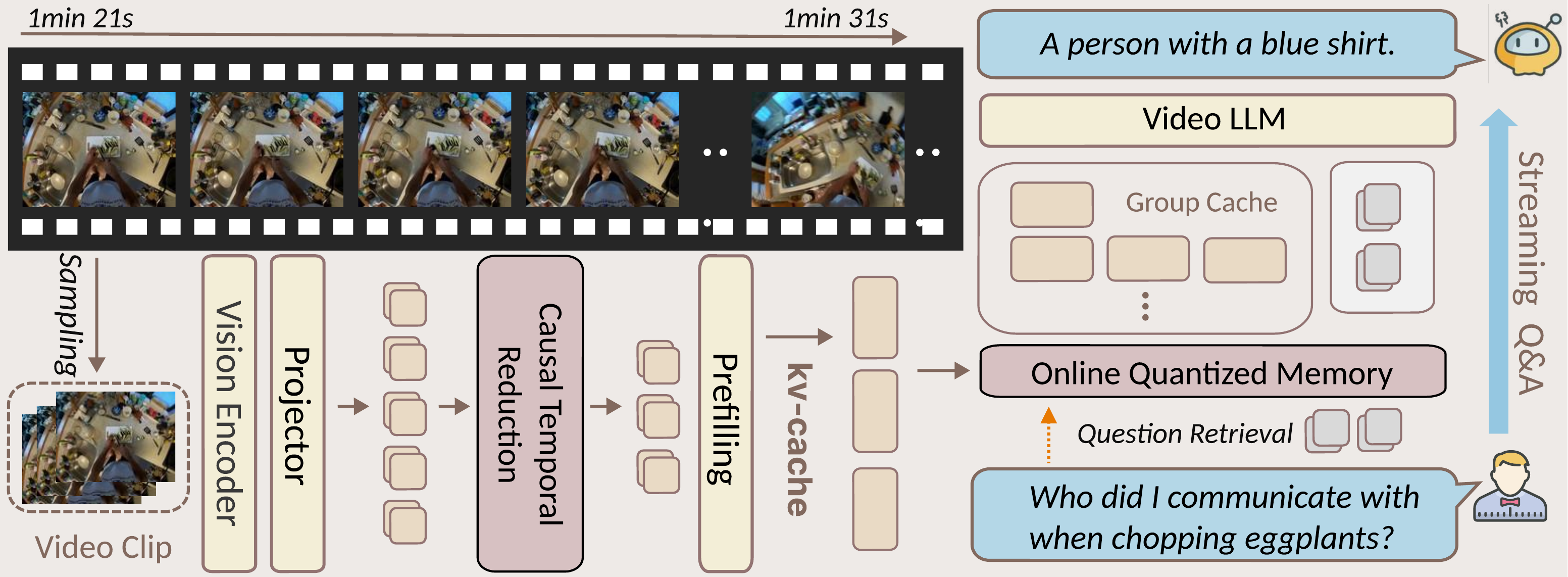} &
\includegraphics[width=0.363\linewidth]{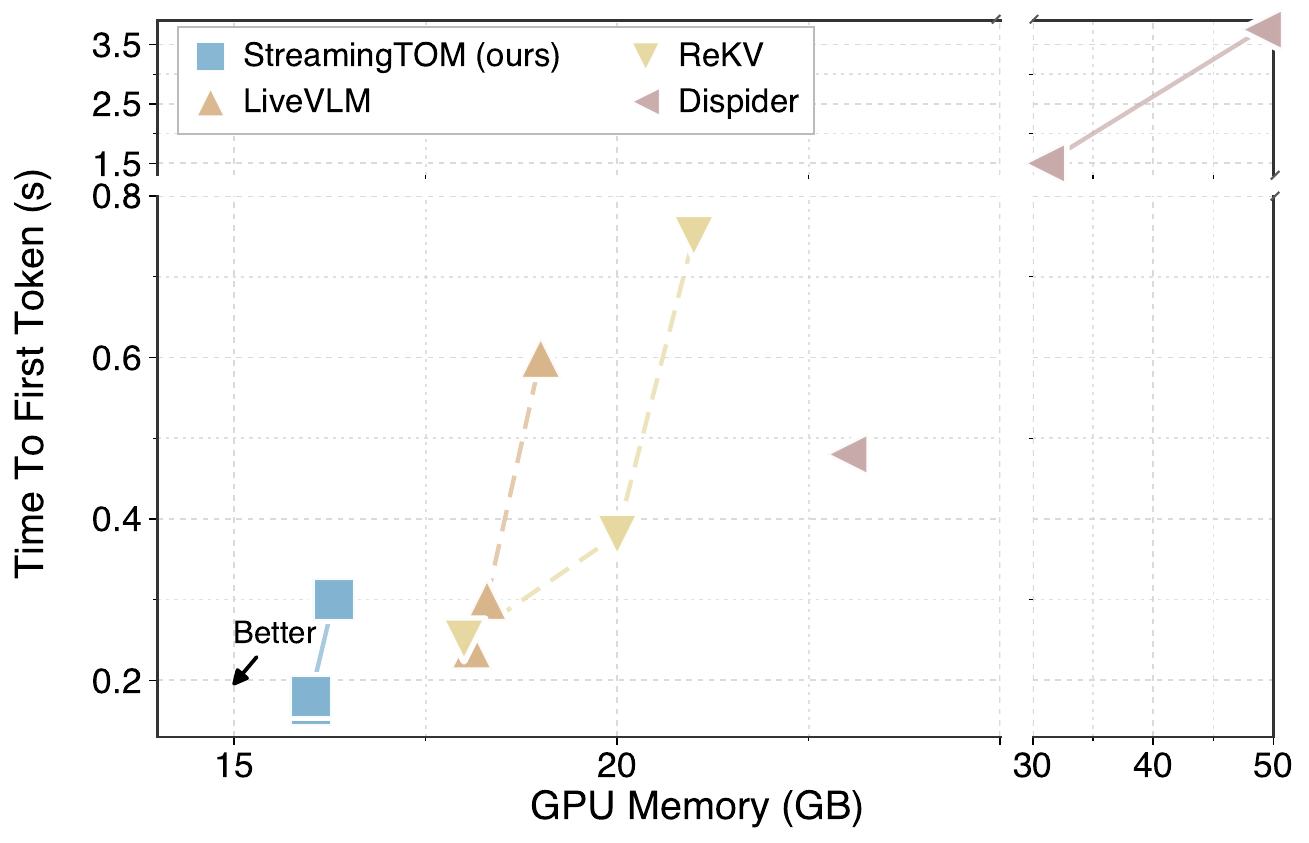}
\end{tabular}
\captionof{figure}{
\textbf{Left}: StreamingTOM (\underline{streaming} \underline{to}ken co\underline{m}pression) is a training-free, two-stage framework for efficient streaming video understanding: \textbf{Causal Temporal Reduction} selects pre-LLM tokens with a causal, fixed-budget policy, and \textbf{Online Quantized Memory} bounds the active kv-cache via 4-bit quantization with on-demand retrieval.
\textbf{Right}: On LLaVA-OV-7B, StreamingTOM achieves a $15.7\times$ kv-cache compression ratio; compared to prior SOTA (LiveVLM), it delivers $1.2\times$ lower peak memory and $2\times$ faster TTFT.
}
\label{fig:teaser}
\end{center}
}]

\begin{abstract}
Unlike offline processing, streaming video vision-language models face two fundamental constraints: causality and accumulation.
Causality prevents access to future frames that offline methods exploit, while accumulation causes tokens to grow unbounded, creating efficiency bottlenecks.
However, existing approaches only regulate post-LLM kv-cache, leaving costly pre-LLM prefill unchanged.
We introduce StreamingTOM, a training-free, plug-and-play two-stage framework that addresses both pre-LLM and post-LLM bottlenecks.
\textbf{Causal Temporal Reduction} imposes a fixed per-frame budget and selects tokens based on adjacent-frame changes and token saliency, drastically reducing per-frame prefill cost by processing only a compact subset of visual tokens, ensuring predictable latency.
\textbf{Online Quantized Memory} stores tokens in 4-bit format, retrieves relevant groups on demand, and dequantizes them, keeping the active kv-cache bounded regardless of stream length.
Experiments demonstrate our method achieves $15.7\times$ kv-cache compression ratio; compared to prior SOTA (LiveVLM), it delivers $1.2\times$ lower peak memory and $2\times$ faster TTFT.
StreamingTOM achieves state-of-the-art accuracy among training-free methods with an average of $63.8\%$ on offline benchmarks and $55.8\%$ accuracy and $3.7$ score on RVS.
These results demonstrate that real-time streaming video understanding with bounded active memory is achievable without model retraining.
\end{abstract}

\section{Introduction}

Recent advances in large language models have significantly improved both reasoning capabilities~\citep{deepseek2025deepseekr1,openai2025o3,kimiteam2026kimik25,qwenteam2025qwq,yang2025qwen3,glm5team2026glm5,llamateam2024llama3,wu2025totrlunlockllmtreeofthoughts,wu2025onpolicyoptimizationgroupequivalent,feng2026rewardmap} and multimodal understanding~\citep{openai2025gpt5,google2026gemini,anthropic2025claude,jin2026mergemix}, demonstrating impressive performance across diverse vision and language tasks.
Building on these foundations, offline video understanding focuses on deep semantic reasoning over prerecorded videos with known temporal boundaries~\citep{bai2023qwenvl,wang2024qwen2vl,bai2025qwen25vl,li2025llava,zhang2025llavavideo,wang2025internvl35}.
A dominant efficiency bottleneck in these models is the computational cost of processing the spatiotemporal redundancy in visual tokens~\citep{jin2025efficientmllm}.
This motivated established compression techniques such as spatiotemporal token merging and dynamic pruning~\citep{shao2026tokens,liu2025revisiting}, which leverage global context, including future frames, to reduce computation with little or no loss in accuracy~\citep{yang2025visionzip,tao2025dycoke,chen2024fastv,shao2025holitom}.
These techniques inherently assume access to a global context and stable clip boundaries.

While offline techniques have matured, emerging applications like autonomous driving, embodied AI, and live video assistants demand streaming processing, where two constraints emerge: causality and accumulation~\citep{lin2024streamingbench,zhang2025flash,wei2025streamvln}.
Causality means the system cannot access future frames.
Accumulation arises because the effective sequence length $L(t)$ grows as frames arrive at a fixed frame rate.
Dense attention grows quadratically with $L(t)$, while the kv-cache grows linearly with time~\citep{ning2025livevlm}.
As compute and memory grow unbounded over time, token compression shifts from an optional optimization to a fundamental prerequisite for streaming viability under fixed budgets.

Contemporary streaming methods either require costly model-specific training~\citep{chen2024videollm,zhang2025flash,xu2026streamingvlm} or operate training-free but focus solely on post-LLM kv-cache management~\citep{di2025streaming,ning2025livevlm,yang2025streammem}.
While the latter reduces memory, it preserves the $O(tNLd^{2})$ pre-LLM computation as all visual tokens must traverse the full transformer stack.
Thus, training-free and strictly causal pre-LLM token reduction, which is essential for true streaming efficiency, remains largely unexplored.

We propose \textbf{StreamingTOM}, a training-free framework that reduces visual tokens before the LLM for streaming video understanding.
Our central insight is that effective streaming compression must act before the LLM under strict causality: post-LLM methods cannot reduce the already-incurred prefill cost, while offline pre-LLM methods violate causality by leveraging future frames.
We impose a fixed per-frame budget $G$ to stabilize latency and ensure predictable compute and memory consumption across all frames.
Accordingly, we implement this design with a strictly causal one-pass selector that makes decisions from adjacent frames $t-1$ and $t$.
\textbf{Causal Temporal Reduction (CTR)} addresses the prefill bottleneck by selecting $G$ tokens per frame based on adjacent-frame changes and token saliency, reducing per-frame prefill cost by a factor of $N/G$ through processing only $G$ tokens instead of all $N$ visual tokens.\footnote{\textbf{Notation:} $N$=tokens per frame, $L$=number of transformer layers, $d$=hidden width, $G$=per-frame retained tokens selected by CTR.}
Meanwhile, \textbf{Online Quantized Memory (OQM)} bounds post-LLM memory growth by storing retained tokens in 4-bit format and retrieving relevant groups on demand, keeping the active kv-cache bounded regardless of stream length.
Together, these two stages enable long-horizon stability under fixed budgets with predictable latency, while remaining plug-and-play across backbones.

StreamingTOM achieves $15.7\times$ kv-cache compression ratio and delivers $1.2\times$ lower peak memory and $2\times$ faster time to first token (TTFT) compared to prior training-free SOTA (LiveVLM~\citep{ning2025livevlm}) in the streaming setting.
It maintains state-of-the-art accuracy among training-free methods with $63.8\%$ average on VideoMME~\citep{fu2025video}, MLVU~\citep{zhou2025mlvu}, and EgoSchema~\citep{mangalam2023egoschema}, and $55.8\%/3.7$ on RVS~\citep{zhang2025flash}.
These gains persist over long horizons because CTR's fixed budget caps per-frame cost while OQM bounds the active kv-cache independently of stream length.
On LLaVA-OV-7B, a one-hour stream reduces kv-cache from $18.8\,\mathrm{GB}$ to $1.2\,\mathrm{GB}$, confirming bounded growth over extended sessions.

In summary, our contributions are threefold.
\begin{itemize}
    \item We introduce strictly causal pre-LLM token reduction for streaming video understanding, reducing prefill complexity from $O(tNLd^{2})$ to $O(tGLd^{2})$ through a fixed per-frame budget $G$ that ensures predictable latency.
    \item We propose StreamingTOM, a training-free two-stage framework unified by a frame-aligned group abstraction of $G$ tokens that coordinates CTR for pre-LLM compression with OQM for bounded post-LLM memory, ensuring temporal coherence across compression and retrieval.
    \item We achieve state-of-the-art training-free accuracy ($63.8\%$ offline average, $55.8\%/3.7$ on RVS) with $15.7\times$ kv-cache compression ratio, $1.2\times$ lower peak memory, and $2\times$ faster TTFT over training-free SOTA, demonstrating both effectiveness and efficiency.
\end{itemize}

\section{Related Work}

\paragraph{Video Token Compression.}
Training-free methods achieve substantial token reduction via spatial selection and adaptive pruning~\citep{bolya2023tome,ren2023testa,yang2025visionzip,xing2025pyramiddrop,huang2025prunevid,shang2025llavaprumerge,chen2025v2drop,yang2025topv}, and dynamic spatio-temporal compression and merging~\citep{shen2025fastvid,tao2025dycoke,tao2025omnizip,liu2025vidcom,liu2026globalcom,shao2025holitom,wang2025dymu}.
However, these approaches rely on global temporal analysis or bidirectional attention that leverages future frames.
This look-ahead capability is incompatible with streaming causality, where compression must depend only on past context and cannot be revised.
Training-based long-video compression methods also exist~\citep{shen2025longvu}, but they require model retraining and are outside our training-free scope.
Beyond token-level reduction, complementary compression strategies such as model pruning~\citep{ma2023llmpruner,zhu2026obsdiff}, kv-cache quantization~\citep{tao2025plugandplay1xbitkvcache,pei2024crossself,wan2024lookm,liu2026mixkv}, and tokenized video compression~\citep{zhou2025tvc} further reduce memory footprint from orthogonal dimensions.

\paragraph{Streaming Video Understanding.}
Training-based methods~\citep{chen2024videollm,zhang2025flash,xu2026streamingvlm,wang2025streambridge,qian2024streaming,qian2025dispider,ding2025streammind,xiong2025streamingvideounderstandingmultiround,fu2025vispeak,chen2025livecc} achieve strong performance through response-timing control, memory decoupling, and perception-decision disentanglement, but require costly model-specific retraining.
TimeChat-Online~\citep{yao2025timechat} observes that 80\% of visual tokens are redundant in streaming videos, motivating aggressive compression.
In contrast, training-free approaches~\citep{di2025streaming,ning2025livevlm,yang2025streammem,kim2025infinipot} focus on post-LLM memory optimization through kv-cache retrieval, eviction, and compression strategies, leaving the full prefill computation cost intact.
Concurrent work~\citep{wang2025stc} also explores token compression for streaming video LLMs.

To our knowledge, no prior training-free streaming method performs strictly causal pre-LLM token reduction.
We address this with StreamingTOM, which couples CTR for pre-LLM compression with OQM for kv-cache management under strict causality.

\begin{figure*}[t!]
    \centering
    \includegraphics[width=1\textwidth]{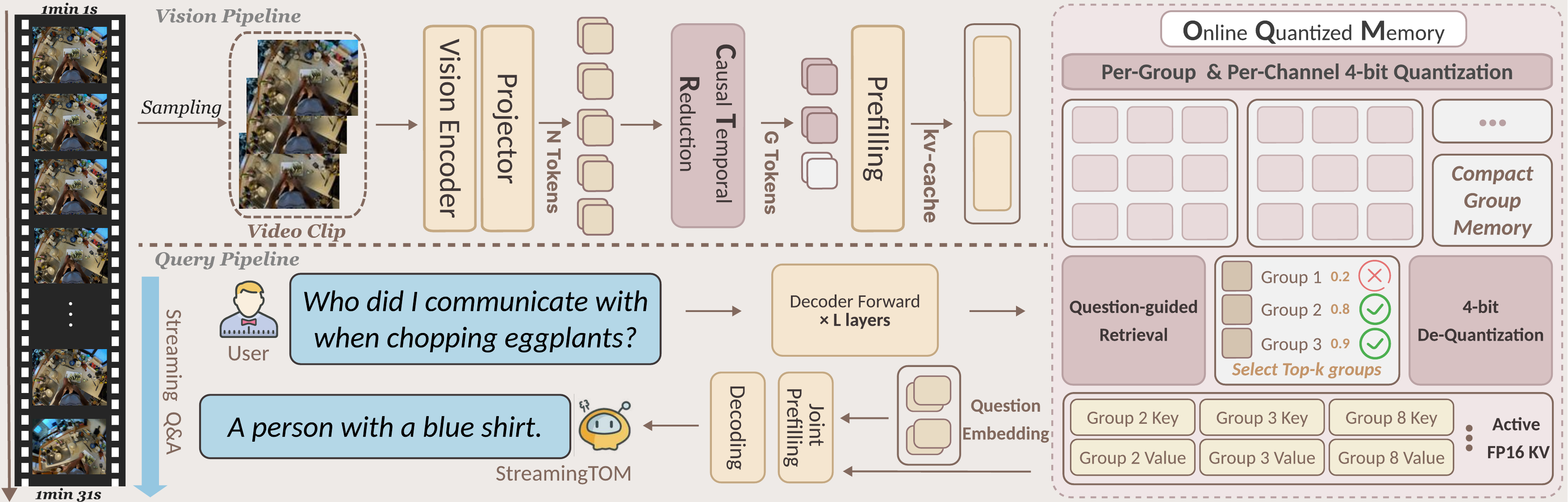}
    \caption{Overview of the StreamingTOM architecture.
    The framework consists of two coordinated pipelines.
    The vision pipeline encodes each frame and applies Causal Temporal Reduction to condense redundant tokens into compact groups, which are written to an online memory for reuse.
    The query pipeline processes questions and drives the decoder to interact with the memory through Online Quantized Memory, which stores groups at 4-bit precision, retrieves at most $k$ groups on demand, and dequantizes them for efficient generation.
    }
    \label{fig:streamingtom_architecture}
\end{figure*}

\section{Methodology}

\subsection{Preliminary}
Streaming video understanding must operate under strict causality when processing an unbounded stream $\mathcal{V}=\{v_1,v_2,\ldots,v_t,\ldots\}$: at time $t$, the system can only access past frames $\mathcal{V}_{\le t}$ while remaining prepared for arbitrary future queries within bounded memory $M$.
This constraint rules out iterative refinement or retrospective reprocessing, since both rely on unavailable future information; instead, each frame must be processed in a single pass and finalized immediately.
In this context, most training-free streaming adaptations of VLMs adopt kv-cache reuse, leveraging the causal transformer decoder to realize the efficient paradigm of ``process once, reuse many''.

Under this paradigm, each incoming frame $v_t$ is encoded once into visual tokens $\mathbf{H}_t=\mathcal{E}_v(v_t)\in\mathbb{R}^{N\times d}$ and, across $L$ transformer layers, mapped to key and value pairs $(K_t^{(l)},V_t^{(l)})$, and the cache accumulates by concatenation $\mathcal{K}^{(l)}_t=[\mathcal{K}^{(l)}_{t-1};K^{(l)}_t]$ and $\mathcal{V}^{(l)}_t=[\mathcal{V}^{(l)}_{t-1};V^{(l)}_t]$.
This creates a fundamental trade-off: single-pass processing avoids repeated work, but because all tokens traverse the full stack, the per-frame compute remains $O(NLd^{2})$; meanwhile, as frames accumulate, the kv-cache grows without bound.
This growth can be formalized as
\begin{equation}
\frac{d\,\text{memory}}{dt}=2\cdot L\cdot N\cdot d\cdot \text{sizeof}(\text{dtype})\cdot \text{fps}.
\label{eq:memory_growth_rate}
\end{equation}
Integrating Eq.~(\ref{eq:memory_growth_rate}) under typical settings yields a total kv-cache footprint of about \textbf{18.8\, GB} for one hour of video\footnote{Based on LLaVA-OV-7B: $L{=}28$, $N{=}196$, $H_{\text{kv}}{=}4$, $d_h{=}128$, FP16, 0.5 fps (1800 frames/hour): $2{\cdot}28{\cdot}1800{\cdot}196{\cdot}4{\cdot}128{\cdot}2\text{B}{\approx}18.8\text{GB}$.}, which substantially exceeds typical GPU memory capacity.

This unbounded growth creates two critical bottlenecks: compute and memory.
Computationally, query processing must attend to all $O(tN)$ accumulated tokens; each decoding step therefore requires $O(tNLd^{2})$ flops, a prohibitive cost for real-time applications.
Post-LLM compression provides no relief because the tokens have already traversed the full transformer stack.
On the memory side, the kv-cache grows linearly over time according to Eq.~(\ref{eq:memory_growth_rate}), quickly exceeding GPU capacity.
Even with advanced eviction strategies, sustaining FP16 precision over long streaming sessions remains infeasible.
Consequently, compression must be performed under a strict, query-agnostic, and causal constraint, optimizing both compute and memory without prior knowledge of the future query distribution.
Let $\mathcal{L}(q, \mathcal{K}, \mathcal{V})$ denote the task loss when answering query $q$ based on the kv-cache $(\mathcal{K}, \mathcal{V})$ derived from the preceding video frames.
The compression objective is then:
\begin{equation}
\min_{(\hat{\mathcal{K}},\hat{\mathcal{V}})\in\mathcal{F}_M}\ \mathbb{E}_{q\sim p}\,\mathcal{L}(q,\hat{\mathcal{K}},\hat{\mathcal{V}}),
\label{eq:compression_objective}
\end{equation}
where $p$ denotes the query distribution, $\mathcal{F}_M = \{(\hat{\mathcal{K}}, \hat{\mathcal{V}}) : |\hat{\mathcal{K}}|+|\hat{\mathcal{V}}|\le M\}$ defines the feasible set for memory budget $M$, and $|\cdot|$ measures memory footprint.
The challenge is to compress to mitigate compute and memory growth while preserving causal validity.

\begin{figure*}[t!]
    \centering
    \includegraphics[width=1\textwidth]{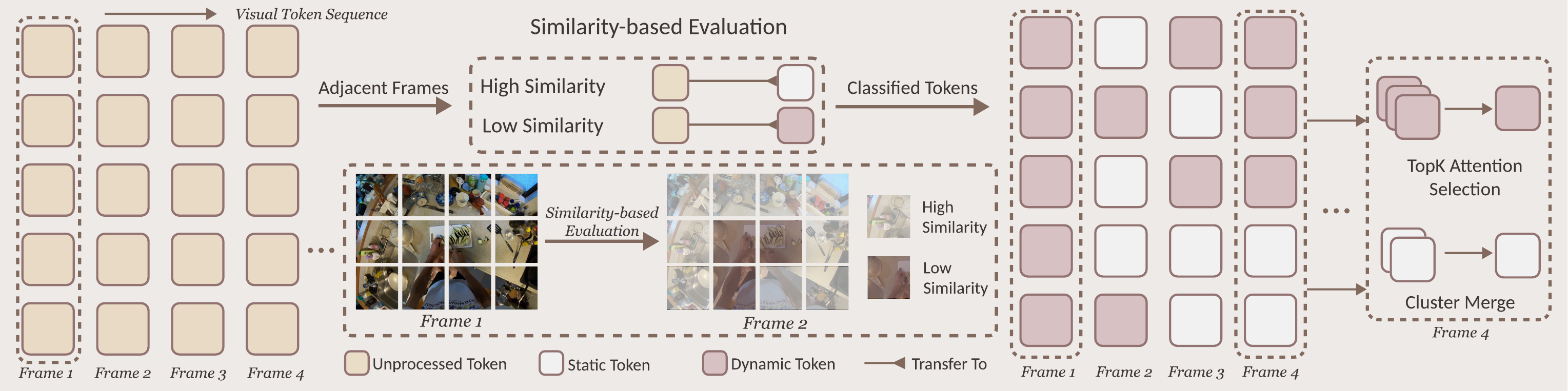}
    \caption{Overview of the CTR compression pipeline.
    The algorithm processes visual tokens from consecutive frames using a 2-frame window (current and previous), producing a binary classification (static, dynamic) through similarity comparison.
    The adaptive budget allocation dynamically distributes compression resources based on content, followed by dual-path processing: dpc clustering for static tokens and attention-based selection for dynamic tokens.}
    \label{fig:compression_algorithm}
\end{figure*}

\subsection{Our Method: StreamingTOM}
The dual bottlenecks identified above demand a coordinated solution, with computational constraints requiring pre-LLM compression and memory constraints requiring post-LLM compression.
Either approach in isolation is insufficient, as pre-LLM compression cannot stop kv-cache accumulation, while post-LLM compression cannot reduce the computational cost already incurred.
The order is also fixed, as tokens must be selected before entering the LLM to reduce computation, and only after LLM processing can they be quantized for storage.

We propose \textbf{StreamingTOM} (\underline{streaming} \underline{to}ken co\underline{m}pression), a unified two-stage framework that integrates Causal Temporal Reduction (CTR) for pre-LLM computational optimization and Online Quantized Memory (OQM) for post-LLM memory management.
As shown in Figure~\ref{fig:streamingtom_architecture}, the two modules are connected through a fixed per-frame budget of $G$ tokens that pins both prefill compute and per-frame kv-writes, making per-frame latency predictable.
Formally, StreamingTOM is defined as the sequential composition of CTR and OQM:
\begin{equation}
\operatorname{StreamingTOM} \;=\; \mathrm{OQM}_{16\to4}\,\circ\,\mathrm{CTR}_{N\to G},
\label{eq:streamingtom_comp}
\end{equation}
where $\mathrm{CTR}_{N\to G}$ reduces per-frame tokens from $N$ to a fixed quota $G$ before the LLM, and $\mathrm{OQM}_{16\to4}$ quantizes stored tensors from FP16 to 4-bit after the LLM while keeping the hidden width $d$ unchanged.

Central to StreamingTOM is the \textbf{group} abstraction, a fixed-size unit of $G$ tokens per frame that serves as the interface between CTR and OQM, decoupling token reduction from storage optimization: CTR selects the most informative $G$ tokens, while OQM handles their efficient storage and retrieval.
This frame-aligned design is critical for streaming: each group corresponds to exactly one frame, ensuring that retrieved content preserves temporal coherence as complete frames rather than fragmented tokens.
Without this coordination, variable-length compression would break frame-aligned retrieval, and token-level retrieval would lose temporal semantics.
The fixed budget $G$ further ensures predictable per-frame latency and enables stable batching across frames with varying content complexity.
Overall, prefill computation reduces from $O(TNLd^{2})$ to $O(TGLd^{2})$, where $T$ denotes the total number of processed frames, and storage reduces from $O(TN\cdot d\cdot 16)$ bits to $O(TG\cdot d\cdot 4)$ bits, resulting in a combined compression ratio of $\frac{4N}{G}$.\footnote{With $N=196$ and $G=50$, this gives $4\times196/50\approx15.7\times$.}

\subsection{Causal Temporal Reduction (CTR)}

CTR addresses these streaming constraints through three design principles: strict causality with a 2-frame window, single-pass processing, and a fixed per-frame budget $G$ for predictable latency.
As illustrated in Figure~\ref{fig:compression_algorithm}, the CTR pipeline evaluates temporal similarity between consecutive frames, classifies tokens into static/dynamic sets, and applies dual-path processing with adaptive budget allocation.

CTR relies on two complementary signals: temporal similarity $s_t^{(i)}$ captures cross-frame redundancy, while spatial saliency $\alpha_t^{(i)}$ identifies informative content within the current frame.
Since the encoder maintains fixed patch indexing across frames, tokens at position $i$ in frames $t$ and $t{-}1$ correspond to the same spatial location.
For temporal similarity, given features $\mathcal{F}_t,\mathcal{F}_{t-1}\!\in\!\mathbb{R}^{N\times d}$, CTR measures per-token consistency by cosine similarity,
\begin{equation}
s_t^{(i)}=\frac{\mathcal{F}_t^{(i)}\cdot\mathcal{F}_{t-1}^{(i)}}{\|\mathcal{F}_t^{(i)}\|\,\|\mathcal{F}_{t-1}^{(i)}\|}.
\end{equation}
For spatial saliency, CTR leverages pre-computed attention-based scores $\alpha_t^{(i)}\!\in\![0,1]$ from the vision encoder as a zero-cost byproduct.
To avoid memory peaks in streaming scenarios, these scores are computed via chunked attention without materializing the full $N{\times}N$ attention matrix.

CTR partitions tokens into disjoint static and dynamic subsets by thresholding the temporal similarity,
\begin{equation}
\mathcal{S}_t = \{i \in [N] \mid s_t^{(i)} > \tau_c\},\qquad \mathcal{D}_t = [N] \setminus \mathcal{S}_t,
\end{equation}
where $\tau_c$ is the similarity threshold.
To compress from $N$ tokens to exactly $G$ tokens per frame, CTR allocates the budget proportionally based on the observed static/dynamic distribution,
\begin{equation}
k_s=\Big\lfloor G\cdot\frac{|\mathcal{S}_t|}{|\mathcal{S}_t|+|\mathcal{D}_t|}\Big\rfloor,\qquad k_d=G-k_s.
\end{equation}
This allocation adapts to content dynamics: allocating more budget to static compression when motion is minimal and prioritizing dynamic selection when changes are salient.
With the budget allocated, CTR applies dual-path processing: for dynamic tokens representing new information, it selects the top-$k_d$ based on saliency scores; for static tokens containing redundant information, it consolidates them through density-based clustering to $k_s$ tokens, yielding
\begin{equation}
\begin{aligned}
\mathcal{G}_d &= \text{TopK}_{k_d}(\{\alpha_t^{(i)}\}_{i\in\mathcal{D}_t}), \\
\mathcal{G}_s &= \text{Merge}_{k_s}(\{\mathcal{F}_t^{(i)}\}_{i\in\mathcal{S}_t}).
\end{aligned}
\end{equation}

The per-frame retained group $\mathcal{G}_t$ combines both paths with $|\mathcal{G}_t|=G$.
This fixed-size output is essential for streaming viability: unlike offline methods, where variable compression ratios are acceptable, streaming systems require predictable per-frame latency and stable batching to maintain real-time guarantees.
CTR maintains only the previous-frame features as state, ensuring decisions are invariant to batch size and enabling consistent results across devices with different memory constraints.
The state memory is $O(Nd)$ independent of stream length, preventing unbounded growth.
Per-frame complexity is $O(N+G^2)$ with $G < N$, dominated by the clustering computation.
This results in a pre-LLM bound that reduces prefill from $O(TNLd^{2})$ to $O(TGLd^{2})$.

\subsection{Online Quantized Memory (OQM)}

Although CTR alleviates computational bottlenecks through pre-LLM compression, memory in the LLM's kv-cache still grows linearly, creating a trade-off between preserving complete history for future queries and operating within bounded capacity.
OQM addresses this by combining quantization with selective retrieval: CTR-produced groups are stored in 4-bit form to reduce storage overhead, while only the groups relevant to a given query are selectively dequantized and activated, enabling complete history preservation with bounded active memory.
This coordination requires a unified operational unit.
The group, the frame-level retained set of exactly $G$ tokens from CTR, serves as the atomic unit for both storage and retrieval, avoiding token-level indexing while preserving frame-aligned semantics.

OQM applies incremental, group-aligned quantization: each incoming group tensor $\mathbf{X}_t \in \mathbb{R}^{H \times G \times d}$ is independently compressed into a 4-bit representation along with a compact retrieval key, without revisiting historical groups,
\begin{equation}
\mathcal{Q}_4(\mathbf{X}_t)=\big(\mathrm{uint4}(\hat{\mathbf{X}}_t),\,\boldsymbol{s}_t,\,\boldsymbol{m}_t,\,\bar{\mathbf{k}}_t\big),
\end{equation}
where the scale and offset are computed per-head, per-channel across the $G$ tokens in each group,
\begin{equation}
\boldsymbol{s}_t = \frac{\max(\mathbf{X}_t)-\min(\mathbf{X}_t)}{15},\qquad \boldsymbol{m}_t = \min(\mathbf{X}_t),
\end{equation}
and a representative key $\bar{\mathbf{k}}_t \in \mathbb{R}^{d}$ is obtained by averaging the keys before quantization.
The 4-bit codes are packed into \texttt{uint8} for storage.
This yields a dual-structure memory $\mathcal{M}_t$: system tokens (instructions and task descriptions) remain in FP16, while quantized visual groups $\mathcal{Q}_4(\mathcal{G}_i)$ with their representative keys $\bar{\mathbf{k}}_i$ are appended incrementally over time.
The per-frame storage ratio from FP16 to 4-bit can be computed as
\begin{equation}
\frac{\text{FP16 storage}}{\text{4-bit storage}} \;\approx\; \frac{N}{G}\cdot\frac{16}{4} \;=\; \frac{4N}{G},
\end{equation}
where the baseline stores $N\!\cdot\!d\!\cdot\!16$ bits and OQM stores $G\!\cdot\!d\!\cdot\!4$ bits per frame.

Only selected groups are restored to full precision via the dequantization operator,
\begin{equation}
\mathcal{Q}_4^{-1}\!\big(\mathrm{uint4}(\hat{\mathbf{X}});\,\boldsymbol{s},\boldsymbol{m}\big)
=\mathrm{depack}(\hat{\mathbf{X}})\odot\boldsymbol{s}+\boldsymbol{m},
\end{equation}
where $\mathrm{depack}(\cdot)$ unpacks 4-bit codes to integers in $\{0,\ldots,15\}$ and $\odot$ denotes elementwise multiplication.
To keep active memory bounded, OQM performs group-level retrieval using the pre-computed representative keys without dequantizing all groups.
For a decoder state $\mathbf{q}$, OQM retrieves the stored representative keys $\bar{\mathbf{k}}_i$ for all groups and computes their similarity to $\mathbf{q}$ in full precision.
It then selects the top-$k$ relevant groups based on cosine similarity,
\begin{equation}
\mathcal{R}=\operatorname{TopK}\big\{\mathrm{sim}(\mathbf{q},\bar{\mathbf{k}}_i)\big\}_{i=1}^{t},
\end{equation}
then dequantizes only the selected groups to form the active kv for decoding,
\begin{equation}
\mathcal{KV}_{\mathrm{active}}=\bigcup_{i\in\mathcal{R}}\,\mathcal{Q}_4^{-1}(\mathcal{G}_i).
\end{equation}
This design enables two critical properties: total storage can scale as $O(T\cdot G\cdot d)/4$ to preserve complete history, while the active kv during decoding remains bounded by $O(k\cdot G\cdot d)$ with $k\!\ll\!T$, ensuring real-time decoding regardless of stream length.
The query-time cost consists of retrieval over $T$ groups plus selective reconstruction $O(k\cdot G)$.
This retrieval-then-dequantize paradigm complements CTR's fixed-budget design: CTR ensures predictable prefill by outputting exactly $G$ tokens per frame, while OQM ensures predictable decoding by retrieving at most $k$ groups, together maintaining bounded latency under streaming causality.
\section{Experiments}

\begin{figure*}[t]
    \centering
    \includegraphics[width=1\linewidth]{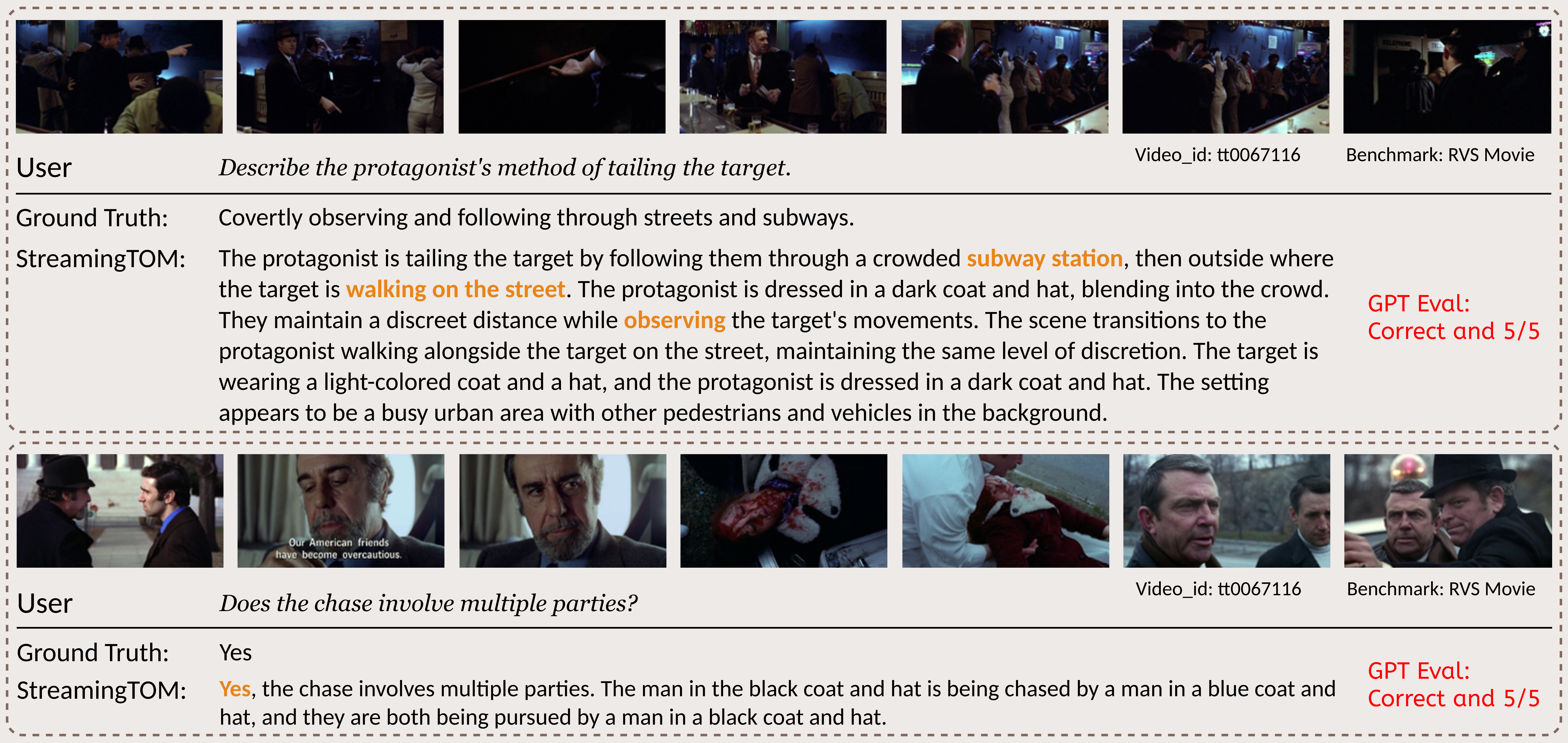}
    \caption{Two qualitative examples from RVS Movie requiring long-horizon reasoning.
    StreamingTOM provides faithful answers consistent with ground truth, accurately capturing fine-grained semantics such as ``subway station'' and multi-party interactions.
    These results illustrate the model's ability to maintain causal reasoning and long-horizon consistency in real streaming scenarios.}
    \label{fig:case_study}
\end{figure*}

\begin{table}
\centering
\caption{Offline evaluation across three long-video benchmarks. ``+'' denotes LLaVA-OV-7B implementations. \textbf{Best} and \underline{second best} in the streaming training-free category.}
\label{tab:main_eval}
\setlength{\tabcolsep}{1mm}
\resizebox{\columnwidth}{!}{%
\begin{tabular}{@{}l c|cccc|cc|c@{}}
\toprule
\multirow{2}{*}{\textbf{Method}} & \multirow{2}{*}{\textbf{Fr.}} & \multicolumn{4}{c|}{\textbf{VideoMME}} & \multirow{2}{*}{\textbf{MLVU}} & \multirow{2}{*}{\textbf{Ego.}} & \multirow{2}{*}{\textbf{Avg.}} \\
\cmidrule(r){3-6}
 & & \textbf{L} & \textbf{M} & \textbf{S} & \textbf{Ovl.} & & & \\
\midrule
\multicolumn{9}{c}{\textit{\textbf{Offline-Designed (Training-free)}}} \\
\midrule
LLaVA-OV-7B \citep{li2025llava} & 32 & 48.8 & 56.4 & 70.1 & 58.4 & 64.7 & 60.1 & 61.0 \\
\textit{+DyCoke} \citep{tao2025dycoke} & 32 & -- & -- & -- & 54.3 & -- & 59.5 & -- \\
\textit{+VisionZip} \citep{yang2025visionzip} & 32 & -- & -- & -- & 58.2 & -- & 60.3 & -- \\
\textit{+HoliTom} \citep{shao2025holitom} & 32 & -- & -- & -- & 58.9 & -- & 61.2 & -- \\
\midrule
\multicolumn{9}{c}{\textit{\textbf{Streaming-Designed (Training-based)}}} \\
\midrule
MovieChat-7B \citep{song2026moviechatplus} & 2048 & 33.4 & -- & -- & 38.2 & 25.8 & 53.5 & 39.2 \\
Dispider-7B \citep{qian2025dispider} & 1fps & 49.7 & 53.7 & -- & 56.5 & 61.7 & 55.6 & 57.9 \\
\midrule
\multicolumn{9}{c}{\textit{\textbf{Streaming-Designed (Training-free)}}} \\
\midrule
LLaVA-OV-7B \citep{li2025llava} & 32 & 48.8 & 56.4 & 70.1 & 58.4 & 64.7 & 60.1 & 61.0 \\
\textit{+ReKV} \citep{di2025streaming} & 0.5fps & -- & -- & -- & -- & \textbf{68.5} & 60.7 & -- \\
\textit{+LiveVLM} \citep{ning2025livevlm} & 0.5/0.2fps & 48.8 & 56.4 & 66.7 & 57.3 & 66.3 & 59.0 & 60.9 \\
\textit{+StreamMem} \citep{yang2025streammem} & 0.5/0.2fps & \underline{50.1} & \underline{56.6} & \textbf{71.5} & \underline{59.4} & 66.9 & \underline{63.0} & 63.1 \\
\textit{\textbf{+StreamingTOM (ours)}} & 0.5/0.2fps & \textbf{50.6} & \textbf{57.8} & \underline{71.3} & \textbf{59.9} & \underline{67.9} & \textbf{63.7} & \textbf{63.8} \\
\bottomrule
\end{tabular}%
}
\end{table}

\subsection{Experimental Setups}

\paragraph{Benchmarks.}
We evaluate StreamingTOM on two tracks: (i) offline long-video understanding and (ii) online streaming qa.
For offline evaluation, we adopt VideoMME~\citep{fu2025video}, MLVU~\citep{zhou2025mlvu}, and EgoSchema~\citep{mangalam2023egoschema}, consistent with streaming methods~\citep{yang2025streammem,ning2025livevlm}.
Videos are sampled at 0.5 frames per second (fps) for clips under 30 minutes and 0.2 fps for longer sequences.
We report VideoMME without subtitles and evaluate MLVU/EgoSchema on official dev splits.
For online streaming, we utilize RVS-Ego and RVS-Movie~\citep{zhang2025flash}, where questions arrive after their end timestamps and require causal reasoning based on preceding content.
These benchmarks are specifically designed to evaluate real-time streaming capabilities under strict causal constraints.

\paragraph{Baselines.}
We compare against four categories of methods.
As the offline reference, we use LLaVA-OV-7B~\citep{li2025llava}, which leverages global temporal access but lacks streaming-specific memory control.
For training-free streaming approaches, we evaluate four methods managing post-LLM kv-cache. ReKV~\citep{di2025streaming} retrieves historical segments from external storage, StreamMem~\citep{yang2025streammem} maintains fixed-size memory through attention-based compression, while LiveVLM~\citep{ning2025livevlm} and InfiniPot-V~\citep{kim2025infinipot} restructure the cache via selective retention and compression.
Additionally, we compare with training-based streaming systems that learn specialized architectures, including Flash-VStream~\citep{zhang2025flash}, Dispider-7B~\citep{qian2025dispider}, and MovieChat-7B~\citep{song2026moviechatplus}.
As offline compression references, we also compare with training-free methods including DyCoke~\citep{tao2025dycoke}, VisionZip~\citep{yang2025visionzip}, and HoliTom~\citep{shao2025holitom}, evaluated on fixed 32-frame clips with 25\% before-LLM token retention~\citep{shao2025holitom}.

\paragraph{Implementation Details.}
We follow the ReKV streaming evaluation protocol~\citep{di2025streaming} and only vary our module hyperparameters; unless noted, prompts and decoding settings follow the corresponding baselines.
All experiments run on a single NVIDIA A6000 (48 GB) GPU using FP16 mixed precision.
We use greedy decoding for deterministic outputs.
For evaluation, long-video benchmarks utilize full conversation templates, while short-video benchmarks employ simplified templates to reduce verbosity.

Pre-LLM reduction with CTR retains 50 tokens per frame based on adjacent-frame changes and token saliency, using a strictly causal two-frame decision without look-ahead.
We set the similarity threshold to 0.9 to determine frame-level redundancy.
Post-LLM memory with OQM stores retained tokens in 4-bit quantized groups and retrieves relevant groups on demand with a total budget of 12k tokens.
For fair comparison with ReKV, we align this budget to match ReKV baseline's $64$ frames at $196$ tokens per frame.
Cached tokens are organized into groups of size 50 to stabilize memory.
The streaming encoder batch size defaults to 32; our compression is batch-agnostic and supports flexible deployment.

\subsection{Main Results}

\begin{table}
  \centering
  \small
  \caption{Evaluation results on RVS streaming benchmarks. Acc denotes accuracy (\%), and Score represents response quality rated on a 1--5 scale. The \textbf{best} result for each metric is in bold.}
  \label{tab:streaming_eval}
  \resizebox{\linewidth}{!}{
  \begin{tabular}{lcccc|cc}
    \toprule
    & \multicolumn{2}{c}{\textbf{RVS-Ego}} & \multicolumn{2}{c}{\textbf{RVS-Movie}} & \multicolumn{2}{c}{\textbf{Avg}} \\
    \cmidrule(lr){2-3} \cmidrule(lr){4-5} \cmidrule(lr){6-7}
    \textbf{Method} & \textbf{Acc} & \textbf{Score} & \textbf{Acc} & \textbf{Score} & \textbf{Acc} & \textbf{Score} \\
    \midrule
    ReKV \citep{di2025streaming} & 63.7 & 4.0 & 54.4 & 3.6 & 59.0 & 3.8 \\
    \midrule
    ReKV w/o off. \citep{di2025streaming} & 55.8 & 3.3 & 50.8 & 3.4 & 53.3 & 3.4 \\
    Flash-VStream \citep{zhang2025flash} & 57.0 & \textbf{4.0} & 53.1 & 3.3 & 55.0 & 3.6 \\
    InfiniPot-V \citep{kim2025infinipot} & 57.9 & 3.5 & 51.4 & 3.5 & 54.6 & 3.5 \\
    StreamMem \citep{yang2025streammem} & 57.6 & 3.8 & 52.7 & 3.4 & 55.2 & 3.6 \\
    \textbf{StreamingTOM (ours)} & \textbf{58.3} & 3.9 & \textbf{53.2} & \textbf{3.5} & \textbf{55.8} & \textbf{3.7} \\
    \bottomrule
  \end{tabular}
  }
\end{table}

\paragraph{Offline Video Evaluation.}

In Table~\ref{tab:main_eval}, StreamingTOM achieves the best overall performance among all streaming systems, regardless of whether they require training.
Our training-free method reaches an average of \textbf{63.8}, surpassing the strongest training-free baseline StreamMem~\citep{yang2025streammem} at 63.1 and outperforming training-based Dispider-7B.
StreamingTOM demonstrates strong performance across individual benchmarks, achieving \textbf{59.9} on VideoMME-Overall, \textbf{67.9} on MLVU, and \textbf{63.7} on EgoSchema.
Within VideoMME splits, StreamingTOM excels particularly on Long and Medium subsets while showing comparable performance on Short videos, demonstrating robust capability across different temporal scales.
These results clearly show that under the same backbone and evaluation protocol, StreamingTOM sets a new SOTA for streaming video understanding despite being entirely training-free.

Beyond streaming comparisons, StreamingTOM surpasses offline compression methods that use 32-frame clips on the same backbone, outperforming HoliTom~\citep{shao2025holitom}, VisionZip~\citep{yang2025visionzip}, and DyCoke~\citep{tao2025dycoke} on VideoMME-Overall and EgoSchema.
This demonstrates that streaming-designed methods, by processing frames causally at 0.5/0.2 fps, adaptively cover longer temporal spans than offline methods constrained to fixed 32-frame clips.  
 
\paragraph{Online Video Evaluation.}

We evaluate StreamingTOM on RVS-Ego and RVS-Movie, using GPT-3.5-turbo-0125~\citep{openai2023gpt35turbo} for grading and limiting GPU memory to 28\, GB without CPU offloading for fair comparison.
Under this memory-constrained setting, StreamingTOM achieves the strongest overall performance with \textbf{55.8 Acc} and \textbf{3.7 Score}, surpassing the strongest training-free baseline StreamMem at 55.2/3.6 and outperforming other methods including InfiniPot-V, ReKV, and training-based Flash-VStream.
While ReKV with CPU offloading reports higher accuracy (59.0), this strategy relies on CPU-GPU memory transfers that are impractical for real-time streaming deployment.
We therefore compare under memory-constrained settings without CPU offloading.
StreamingTOM achieves \textbf{58.3}/\textbf{3.9} on RVS-Ego and \textbf{53.2}/\textbf{3.5} on RVS-Movie, demonstrating consistent performance across both.
We further provide qualitative examples on RVS Movie (Figure~\ref{fig:case_study}), where StreamingTOM produces faithful answers aligned with ground truth, demonstrating its ability to capture fine-grained semantics under causal streaming constraints.

\subsection{Efficiency Analysis}

\begin{figure*}[t!]
  \centering
  \includegraphics[width=1\linewidth]{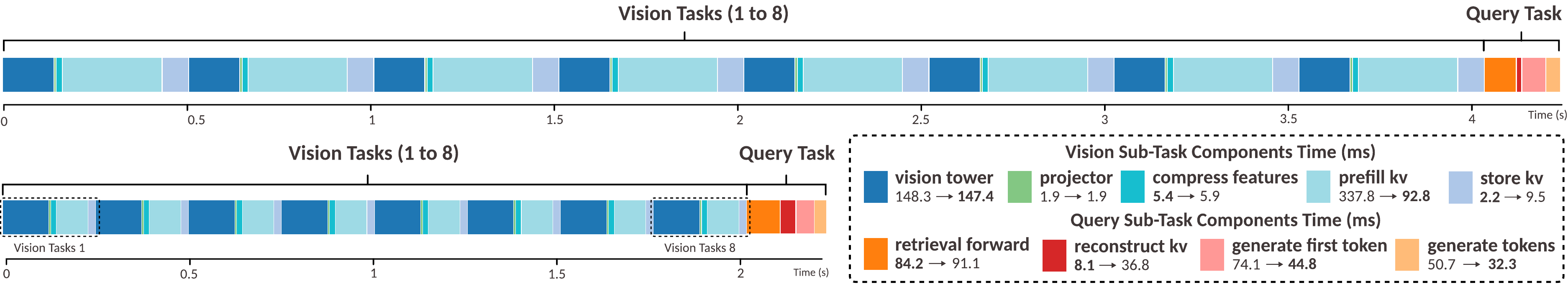}
  \caption{\textbf{Complete processing timeline comparison.} The top row shows the baseline, while the bottom row shows StreamingTOM.
  For a 64-frame stream at batch size 8 with 50 tokens per frame, StreamingTOM incurs minimal overhead: OQM introduces 7.3ms for kv storage, 6.9ms for retrieval, and 28.7ms for 4-bit reconstruction.
  In contrast, CTR delivers 3.6$\times$ prefill acceleration from 337.8ms to 92.8ms, yielding an efficient 0.20s query TTFT and demonstrating substantial performance gains at negligible cost.
  }
  \label{fig:eff_breakdown}
\end{figure*}

\begin{table}
\centering
\caption{StreamingTOM efficiency metrics across frame budgets and sampling rates under two batch sizes. Bounded memory growth and consistent throughput confirm real-time capability.}
\label{tab:efficiency_centered}
\resizebox{\columnwidth}{!}{%
\begin{tabular}{l cccc|cc}
\toprule
 & \multicolumn{4}{c|}{\textbf{Frames}} & \multicolumn{2}{c}{\textbf{Streaming}} \\
\cmidrule(lr){2-5} \cmidrule(lr){6-7}
\textbf{Metric} & \textbf{16} & \textbf{64} & \textbf{256} & \textbf{512} & \textbf{0.2 fps} & \textbf{0.5 fps} \\
\midrule
\multicolumn{7}{c}{\textit{\textbf{Batch Size: 8}}} \\
GPU Mem. / GB $\downarrow$          & 16.0 & 16.0 & 16.3 & 16.7 & 16.9 & 18.4 \\
TTFT / s $\downarrow$               & 0.17 & 0.20 & 0.30 & 0.30 & 0.31 & 0.36 \\
Throughput / tokens$\cdot$s$^{-1}$ $\uparrow$    & 36.7 & 32.6 & 20.8 & 20.9 & 20.8 & 20.9 \\
\midrule
\multicolumn{7}{c}{\textit{\textbf{Batch Size: 32}}} \\
GPU Mem. / GB $\downarrow$          & 16.7 & 18.6 & 18.9 & 19.3 & 19.5 & 21.1 \\
TTFT / s $\downarrow$               & 0.18 & 0.18 & 0.28 & 0.29 & 0.29 & 0.30 \\
Throughput / tokens$\cdot$s$^{-1}$ $\uparrow$    & 36.5 & 32.4 & 20.9 & 20.9 & 21.0 & 20.9 \\
\bottomrule
\end{tabular}%
}
\end{table}

\begin{figure}[t!]
    \centering
    \includegraphics[width=1\linewidth]{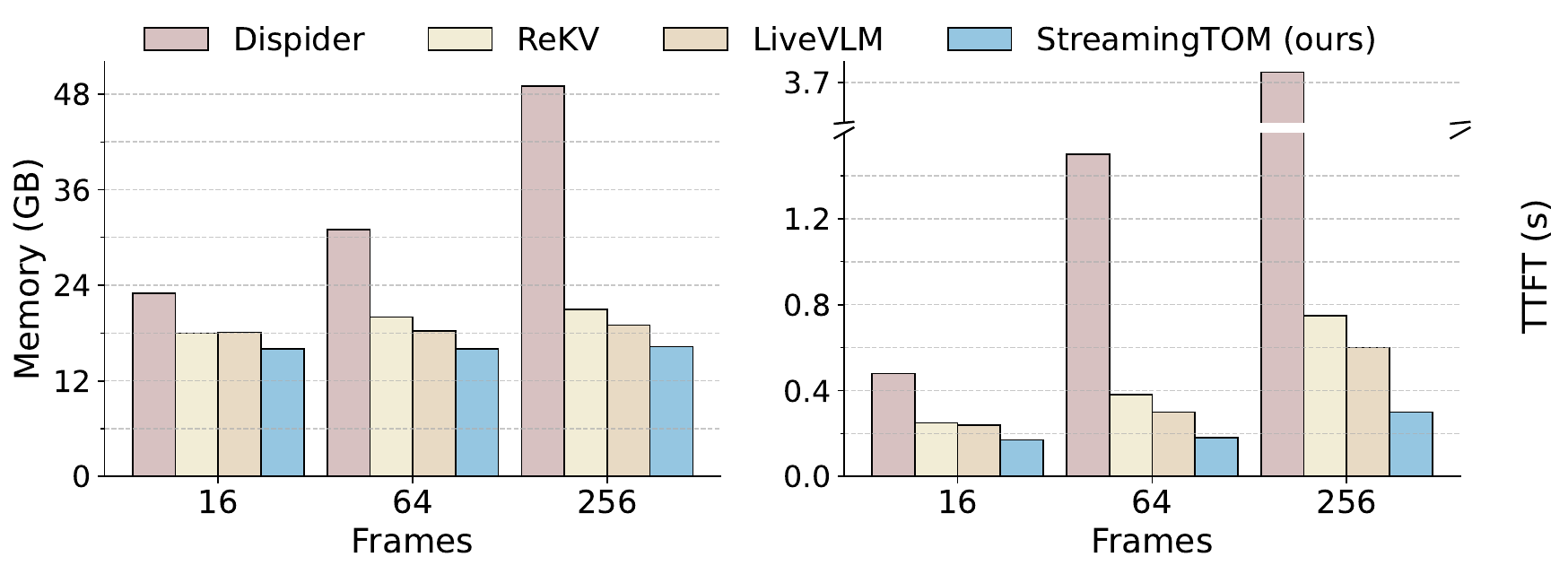}
    \caption{Memory and TTFT comparison across frame budgets. StreamingTOM outperforms all baselines in both metrics.}
    \label{fig:eff_memory}
\end{figure}

We evaluate efficiency using three metrics (peak memory, TTFT, and generation throughput) across varying frame budgets, sampling rates, and batch sizes.
Our compression is batch-agnostic, maintaining consistent accuracy across all batch sizes.
Figure~\ref{fig:eff_memory} compares StreamingTOM in detail with representative training-free baselines.
Unlike LiveVLM, which compresses kv-cache post-LLM, our dual-stage design achieves superior efficiency: \textbf{CTR} reduces visual tokens \emph{before} the LLM, cutting the $O(tNLd^{2})$ prefill computation and kv-writes, while \textbf{OQM} further compresses the resulting kv-cache through 4-bit quantization and bounded retrieval.
StreamingTOM exhibits significantly flatter memory growth due to a smaller and more stable active kv-cache, with advantages widening as the number of frames increases: at 256 frames, StreamingTOM achieves $1.2\times$ lower peak memory and $2\times$ faster TTFT compared to LiveVLM.

Table~\ref{tab:efficiency_centered} provides a comprehensive performance profile of StreamingTOM across different configurations to guide practical deployment.
Memory exhibits sublinear growth consistently due to CTR's fixed token budget, remaining nearly constant from 16.0 GB to 16.7 GB across 16-512 frames, demonstrating bounded memory under pre-LLM compression.
TTFT remains consistently low and throughput stabilizes at approximately 20 tokens/s for long sequences across varying video lengths and batch sizes, confirming scalability for practical streaming deployment.
Figure~\ref{fig:eff_breakdown} breaks down the complete processing pipeline, comparing baseline and StreamingTOM across vision and query stages.
These results confirm that StreamingTOM maintains efficient and predictable performance across diverse streaming scenarios.

\subsection{Ablation Study}

\begin{table}
\centering
\caption{Ablation on token retention and quantization. Ratio is relative to a 196-token, 16-bit baseline. $^*$Results for 50 tokens with 2-bit quantization are interpolated from 48 and 52 tokens.}
\label{tab:ablation_token_quant}
\resizebox{\columnwidth}{!}{%
\begin{tabular}{c c c c c c c}
\toprule
\multirow{2}{*}{\textbf{Token}} & \multirow{2}{*}{\textbf{Quant.}} & \multirow{2}{*}{\textbf{Ratio (\%)}} & \multicolumn{4}{c}{\textbf{VideoMME}$\uparrow$} \\
& & & \textbf{Short} & \textbf{Medium} & \textbf{Long} & \textbf{Overall} \\
\midrule
\multirow{2}{*}{40} & 4-bit & 5.1 & 70.1 & 56.8 & 49.9 & 58.9 \\
                    & 2-bit & 2.6 & 69.1 & 57.1 & 48.0 & 58.1 \\
\midrule
\multirow{2}{*}{50} & 4-bit & 6.4 & \textbf{71.3} & \textbf{57.8} & \textbf{50.6} & \textbf{59.9} \\
                    & 2-bit$^*$ & 3.2 & 69.3 & 57.3 & 48.8 & 58.5 \\
\midrule
\multirow{2}{*}{60} & 4-bit & 7.7 & 69.9 & 57.6 & 50.4 & 59.3 \\
                    & 2-bit & 3.8 & 68.9 & 56.8 & 50.3 & 58.7 \\
\bottomrule
\end{tabular}%
}
\end{table}

Table~\ref{tab:ablation_token_quant} analyzes the trade-off between per-frame detail and temporal coverage.
Our default configuration (50 tokens, 4-bit) achieves the best balance: fewer tokens (40) sacrifice critical details, while more tokens (60) reduce temporal coverage under fixed memory budgets, both degrading accuracy.
This configuration reduces memory footprint to 6.4\% relative to the 196-token 16-bit baseline, achieving a $15.7\times$ compression ratio while maintaining strong accuracy.
While 2-bit quantization further halves memory, it consistently yields lower accuracy across all token counts, confirming that 4-bit strikes the optimal balance between compression and fidelity.
\section{Conclusion}

This paper presents StreamingTOM, the first training-free framework to perform pre-LLM token reduction for streaming video understanding.
Unlike existing methods that only manage post-LLM kv-cache, StreamingTOM addresses pre-LLM computation and post-LLM memory bottlenecks through a unified two-stage design.
CTR enforces strict causality with a fixed per-frame budget, reducing prefill complexity, while OQM bounds post-LLM memory through 4-bit quantization and retrieval.
Experiments demonstrate that StreamingTOM achieves state-of-the-art accuracy among training-free methods, reaching 63.8\% on offline benchmarks and 55.8\%/3.7 on RVS with $15.7\times$ kv-cache compression ratio.
Critically, it sustains bounded memory growth over long horizons by reducing one-hour stream kv-cache from 18.8GB to 1.2GB, enabling deployment of video LLMs for real-time applications.

\section*{Acknowledgments}
This paper is supported by Young Scientists Fund of the National Natural Science Foundation of China (NSFC) (No. 62506305), Zhejiang Leading Innovative and Entrepreneur Team Introduction Program (No. 2024R01007), Key Research and Development Program of Zhejiang Province (No. 2025C01026), Scientific Research Project of Westlake University (No. WU2025WF003), Chinese Association for Artificial Intelligence (CAAI) \& Ant Group Research Fund - AGI Track (No. 2025CAAI-ANT-13). It is also supported by the research funds of the National Talent Program and Hangzhou Municipal Talent Program.

{
    \small
    \bibliographystyle{ieeenat_fullname}
    \bibliography{main}

\begin{thebibliography}{74}
\providecommand{\natexlab}[1]{#1}
\providecommand{\url}[1]{\texttt{#1}}
\expandafter\ifx\csname urlstyle\endcsname\relax
  \providecommand{\doi}[1]{doi: #1}\else
  \providecommand{\doi}{doi: \begingroup \urlstyle{rm}\Url}\fi

\bibitem[{Anthropic}(2025)]{anthropic2025claude}
{Anthropic}.
\newblock Claude sonnet 4.5.
\newblock \url{https://www.anthropic.com/news/claude-sonnet-4-5}, 2025.

\bibitem[Bai et~al.(2023)Bai, Bai, Yang, Wang, Tan, Wang, Lin, Zhou, and Zhou]{bai2023qwenvl}
Jinze Bai, Shuai Bai, Shusheng Yang, Shijie Wang, Sinan Tan, Peng Wang, Junyang Lin, Chang Zhou, and Jingren Zhou.
\newblock Qwen-vl: A versatile vision-language model for understanding, localization, text reading, and beyond.
\newblock \emph{arXiv preprint arXiv:2308.12966}, 2023.

\bibitem[Bai et~al.(2025)Bai, Chen, Liu, Wang, Ge, Song, Dang, Wang, Wang, Tang, Zhong, Zhu, Yang, Li, Wan, Wang, Ding, Fu, Xu, Ye, Zhang, Xie, Cheng, Zhang, Yang, Xu, and Lin]{bai2025qwen25vl}
Shuai Bai, Keqin Chen, Xuejing Liu, Jialin Wang, Wenbin Ge, Sibo Song, Kai Dang, Peng Wang, Shijie Wang, Jun Tang, Humen Zhong, Yuanzhi Zhu, Mingkun Yang, Zhaohai Li, Jianqiang Wan, Pengfei Wang, Wei Ding, Zheren Fu, Yiheng Xu, Jiabo Ye, Xi Zhang, Tianbao Xie, Zesen Cheng, Hang Zhang, Zhibo Yang, Haiyang Xu, and Junyang Lin.
\newblock Qwen2.5-vl technical report.
\newblock \emph{arXiv preprint arXiv:2502.13923}, 2025.

\bibitem[Bolya et~al.(2023)Bolya, Fu, Dai, Zhang, Feichtenhofer, and Hoffman]{bolya2023tome}
Daniel Bolya, Cheng-Yang Fu, Xiaoliang Dai, Peizhao Zhang, Christoph Feichtenhofer, and Judy Hoffman.
\newblock Token merging: Your vit but faster.
\newblock In \emph{ICLR}, 2023.

\bibitem[Chen et~al.(2024{\natexlab{a}})Chen, Lv, Wu, Lin, Song, Gao, Liu, Gao, Mao, and Shou]{chen2024videollm}
Joya Chen, Zhaoyang Lv, Shiwei Wu, Kevin~Qinghong Lin, Chenan Song, Difei Gao, Jia-Wei Liu, Ziteng Gao, Dongxing Mao, and Mike~Zheng Shou.
\newblock Videollm-online: Online video large language model for streaming video.
\newblock In \emph{CVPR}, 2024{\natexlab{a}}.

\bibitem[Chen et~al.(2025{\natexlab{a}})Chen, Liu, Wen, Wang, Huang, and Chen]{chen2025v2drop}
Junjie Chen, Xuyang Liu, Zichen Wen, Yiyu Wang, Siteng Huang, and Honggang Chen.
\newblock Variation-aware vision token dropping for faster large vision-language models.
\newblock \emph{arXiv preprint arXiv:2509.01552}, 2025{\natexlab{a}}.

\bibitem[Chen et~al.(2025{\natexlab{b}})Chen, Zeng, Lin, Li, Ma, and Shou]{chen2025livecc}
Joya Chen, Ziyun Zeng, Yiqi Lin, Wei Li, Zejun Ma, and Mike~Zheng Shou.
\newblock Livecc: Learning video llm with streaming speech transcription at scale.
\newblock In \emph{CVPR}, 2025{\natexlab{b}}.

\bibitem[Chen et~al.(2024{\natexlab{b}})Chen, Zhao, Liu, Bai, Lin, Zhou, and Chang]{chen2024fastv}
Liang Chen, Haozhe Zhao, Tianyu Liu, Shuai Bai, Junyang Lin, Chang Zhou, and Baobao Chang.
\newblock An image is worth 1/2 tokens after layer 2: Plug-and-play inference acceleration for large vision-language models.
\newblock In \emph{ECCV}, 2024{\natexlab{b}}.

\bibitem[Dao(2024)]{dao2024flashattention2}
Tri Dao.
\newblock Flashattention-2: Faster attention with better parallelism and work partitioning.
\newblock In \emph{ICLR}, 2024.

\bibitem[{DeepSeek-AI}(2025)]{deepseek2025deepseekr1}
{DeepSeek-AI}.
\newblock Deepseek-r1 incentivizes reasoning in llms through reinforcement learning.
\newblock \emph{Nature}, 645:\penalty0 633--638, 2025.

\bibitem[Di et~al.(2025)Di, Yu, Zhang, Li, Zhong, Cheng, Li, He, Shu, and Jiang]{di2025streaming}
Shangzhe Di, Zhelun Yu, Guanghao Zhang, Haoyuan Li, Tao Zhong, Hao Cheng, Bolin Li, Wanggui He, Fangxun Shu, and Hao Jiang.
\newblock Streaming video question-answering with in-context video kv-cache retrieval.
\newblock In \emph{ICLR}, 2025.

\bibitem[Ding et~al.(2025)Ding, Wu, Yang, Jiang, Zhang, Bai, Chen, and Cao]{ding2025streammind}
Xin Ding, Hao Wu, Yifan Yang, Shiqi Jiang, Qianxi Zhang, Donglin Bai, Zhibo Chen, and Ting Cao.
\newblock Streammind: Unlocking full frame rate streaming video dialogue through event-gated cognition.
\newblock In \emph{ICCV}, 2025.

\bibitem[Feng et~al.(2026)Feng, Tuo, Wang, Kong, Zhu, and Wang]{feng2026rewardmap}
Sicheng Feng, Kaiwen Tuo, Song Wang, Lingdong Kong, Jianke Zhu, and Huan Wang.
\newblock Rewardmap: Tackling sparse rewards in fine-grained visual reasoning via multi-stage reinforcement learning.
\newblock In \emph{ICLR}, 2026.

\bibitem[Fu et~al.(2025{\natexlab{a}})Fu, Dai, Luo, Li, Ren, Zhang, Wang, Zhou, Shen, Zhang, Chen, Li, Lin, Zhao, Li, Xu, Zheng, Chen, Shan, He, and Sun]{fu2025video}
Chaoyou Fu, Yuhan Dai, Yongdong Luo, Lei Li, Shuhuai Ren, Renrui Zhang, Zihan Wang, Chenyu Zhou, Yunhang Shen, Mengdan Zhang, Peixian Chen, Yanwei Li, Shaohui Lin, Sirui Zhao, Ke Li, Tong Xu, Xiawu Zheng, Enhong Chen, Caifeng Shan, Ran He, and Xing Sun.
\newblock Video-mme: The first-ever comprehensive evaluation benchmark of multi-modal llms in video analysis.
\newblock In \emph{CVPR}, 2025{\natexlab{a}}.

\bibitem[Fu et~al.(2025{\natexlab{b}})Fu, Yang, Li, Peng, Lin, Wei, Hu, Xie, and Zheng]{fu2025vispeak}
Shenghao Fu, Qize Yang, Yuan-Ming Li, Yi-Xing Peng, Kun-Yu Lin, Xihan Wei, Jian-Fang Hu, Xiaohua Xie, and Wei-Shi Zheng.
\newblock Vispeak: Visual instruction feedback in streaming videos.
\newblock In \emph{ICCV}, 2025{\natexlab{b}}.

\bibitem[{GLM-5 Team}(2026)]{glm5team2026glm5}
{GLM-5 Team}.
\newblock Glm-5: from vibe coding to agentic engineering.
\newblock \emph{arXiv preprint arXiv:2602.15763}, 2026.

\bibitem[{Google DeepMind}(2026)]{google2026gemini}
{Google DeepMind}.
\newblock Gemini 3.1 pro.
\newblock \url{https://deepmind.google/models/gemini/pro/}, 2026.

\bibitem[Huang et~al.(2025)Huang, Zhou, and Han]{huang2025prunevid}
Xiaohu Huang, Hao Zhou, and Kai Han.
\newblock Prunevid: Visual token pruning for efficient video large language models.
\newblock In \emph{Findings of ACL}, 2025.

\bibitem[Jin et~al.(2026)Jin, Li, Jian, Yu, and Wang]{jin2026mergemix}
Xin Jin, Siyuan Li, Siyong Jian, Kai Yu, and Huan Wang.
\newblock Mergemix: A unified augmentation paradigm for visual and multi-modal understanding.
\newblock In \emph{ICLR}, 2026.

\bibitem[Jin et~al.(2025)Jin, Li, Gu, Liu, Zhao, Lai, Gan, Wang, Wang, Tan, and Ma]{jin2025efficientmllm}
Yizhang Jin, Jian Li, Tianjun Gu, Yexin Liu, Bo Zhao, Jinxiang Lai, Zhenye Gan, Yabiao Wang, Chengjie Wang, Xin Tan, and Lizhuang Ma.
\newblock Efficient multimodal large language models: A survey.
\newblock \emph{Visual Intelligence}, 3:\penalty0 27, 2025.

\bibitem[Kim et~al.(2025)Kim, Shim, Choi, and Chang]{kim2025infinipot}
Minsoo Kim, Kyuhong Shim, Jungwook Choi, and Simyung Chang.
\newblock Infinipot-v: Memory-constrained kv cache compression for streaming video understanding.
\newblock In \emph{NeurIPS}, 2025.

\bibitem[{Kimi Team}(2026)]{kimiteam2026kimik25}
{Kimi Team}.
\newblock Kimi k2.5: Visual agentic intelligence.
\newblock \emph{arXiv preprint arXiv:2602.02276}, 2026.

\bibitem[Li et~al.(2025)Li, Zhang, Guo, Zhang, Li, Zhang, Zhang, Zhang, Li, Liu, and Li]{li2025llava}
Bo Li, Yuanhan Zhang, Dong Guo, Renrui Zhang, Feng Li, Hao Zhang, Kaichen Zhang, Peiyuan Zhang, Yanwei Li, Ziwei Liu, and Chunyuan Li.
\newblock Llava-onevision: Easy visual task transfer.
\newblock \emph{TMLR}, 2025.

\bibitem[Lin et~al.(2024)Lin, Fang, Chen, Wan, Luo, Li, Liu, and Sun]{lin2024streamingbench}
Junming Lin, Zheng Fang, Chi Chen, Zihao Wan, Fuwen Luo, Peng Li, Yang Liu, and Maosong Sun.
\newblock Streamingbench: Assessing the gap for mllms to achieve streaming video understanding.
\newblock \emph{arXiv preprint arXiv:2411.03628}, 2024.

\bibitem[Liu et~al.(2025{\natexlab{a}})Liu, Lin, Wei, Shao, Tao, Huang, Yang, Chen, Wang, and Jin]{liu2025revisiting}
Jinming Liu, Junyan Lin, Yuntao Wei, Kele Shao, Keda Tao, Jianguo Huang, Xudong Yang, Zhibo Chen, Huan Wang, and Xin Jin.
\newblock Revisiting mllm token technology through the lens of classical visual coding.
\newblock \emph{arXiv preprint arXiv:2508.13460}, 2025{\natexlab{a}}.

\bibitem[Liu et~al.(2025{\natexlab{b}})Liu, Wang, Ma, and Zhang]{liu2025vidcom}
Xuyang Liu, Yiyu Wang, Junpeng Ma, and Linfeng Zhang.
\newblock Video compression commander: Plug-and-play inference acceleration for video large language models.
\newblock In \emph{EMNLP}, 2025{\natexlab{b}}.

\bibitem[Liu et~al.(2026{\natexlab{a}})Liu, Gui, Zhang, and Zhang]{liu2026mixkv}
Xuyang Liu, Xiyan Gui, Yuchao Zhang, and Linfeng Zhang.
\newblock Mixing importance with diversity: Joint optimization for kv cache compression in large vision-language models.
\newblock In \emph{ICLR}, 2026{\natexlab{a}}.

\bibitem[Liu et~al.(2026{\natexlab{b}})Liu, Wang, Chen, Han, Wang, Yuan, Song, Huang, and Chen]{liu2026globalcom}
Xuyang Liu, Ziming Wang, Junjie Chen, Yuhang Han, Yingyao Wang, Jiale Yuan, Jun Song, Siteng Huang, and Honggang Chen.
\newblock Global compression commander: Plug-and-play inference acceleration for high-resolution large vision-language models.
\newblock In \emph{AAAI}, 2026{\natexlab{b}}.

\bibitem[{Llama Team, AI @ Meta}(2024)]{llamateam2024llama3}
{Llama Team, AI @ Meta}.
\newblock The llama 3 herd of models.
\newblock \emph{arXiv preprint arXiv:2407.21783}, 2024.

\bibitem[Ma et~al.(2023)Ma, Fang, and Wang]{ma2023llmpruner}
Xinyin Ma, Gongfan Fang, and Xinchao Wang.
\newblock Llm-pruner: On the structural pruning of large language models.
\newblock In \emph{NeurIPS}, 2023.

\bibitem[Mangalam et~al.(2023)Mangalam, Akshulakov, and Malik]{mangalam2023egoschema}
Karttikeya Mangalam, Raiymbek Akshulakov, and Jitendra Malik.
\newblock Egoschema: A diagnostic benchmark for very long-form video language understanding.
\newblock In \emph{NeurIPS}, 2023.

\bibitem[Ning et~al.(2025)Ning, Liu, Jin, Ding, Guo, and Zhao]{ning2025livevlm}
Zhenyu Ning, Guangda Liu, Qihao Jin, Wenchao Ding, Minyi Guo, and Jieru Zhao.
\newblock Livevlm: Efficient online video understanding via streaming-oriented kv cache and retrieval.
\newblock \emph{arXiv preprint arXiv:2505.15269}, 2025.

\bibitem[{OpenAI}(2023)]{openai2023gpt35turbo}
{OpenAI}.
\newblock Gpt-3.5 turbo.
\newblock \url{https://platform.openai.com/docs/models/gpt-3.5-turbo}, 2023.

\bibitem[{OpenAI}(2025{\natexlab{a}})]{openai2025gpt5}
{OpenAI}.
\newblock Gpt-5.
\newblock \url{https://openai.com/gpt-5/}, 2025{\natexlab{a}}.

\bibitem[{OpenAI}(2025{\natexlab{b}})]{openai2025o3}
{OpenAI}.
\newblock Introducing openai o3 and o4-mini.
\newblock \url{https://openai.com/index/introducing-o3-and-o4-mini/}, 2025{\natexlab{b}}.

\bibitem[Pei et~al.(2024)Pei, Huang, and Xu]{pei2024crossself}
Xiaohuan Pei, Tao Huang, and Chang Xu.
\newblock Cross-self kv cache pruning for efficient vision-language inference.
\newblock \emph{arXiv preprint arXiv:2412.04652}, 2024.

\bibitem[Qian et~al.(2024)Qian, Dong, Zhang, Zang, Ding, Lin, and Wang]{qian2024streaming}
Rui Qian, Xiaoyi Dong, Pan Zhang, Yuhang Zang, Shuangrui Ding, Dahua Lin, and Jiaqi Wang.
\newblock Streaming long video understanding with large language models.
\newblock In \emph{NeurIPS}, 2024.

\bibitem[Qian et~al.(2025)Qian, Ding, Dong, Zhang, Zang, Cao, Lin, and Wang]{qian2025dispider}
Rui Qian, Shuangrui Ding, Xiaoyi Dong, Pan Zhang, Yuhang Zang, Yuhang Cao, Dahua Lin, and Jiaqi Wang.
\newblock Dispider: Enabling video llms with active real-time interaction via disentangled perception, decision, and reaction.
\newblock In \emph{CVPR}, 2025.

\bibitem[{Qwen Team}(2025)]{qwenteam2025qwq}
{Qwen Team}.
\newblock Qwq-32b: Embracing the power of reinforcement learning.
\newblock \url{https://qwenlm.github.io/blog/qwq-32b/}, 2025.

\bibitem[Ren et~al.(2023)Ren, Chen, Li, Sun, and Hou]{ren2023testa}
Shuhuai Ren, Sishuo Chen, Shicheng Li, Xu Sun, and Lu Hou.
\newblock Testa: Temporal-spatial token aggregation for long-form video-language understanding.
\newblock In \emph{Findings of EMNLP}, 2023.

\bibitem[Shang et~al.(2025)Shang, Cai, Xu, Lee, and Yan]{shang2025llavaprumerge}
Yuzhang Shang, Mu Cai, Bingxin Xu, Yong~Jae Lee, and Yan Yan.
\newblock Llava-prumerge: Adaptive token reduction for efficient large multimodal models.
\newblock In \emph{ICCV}, 2025.

\bibitem[Shao et~al.(2025)Shao, Tao, Qin, You, Sui, and Wang]{shao2025holitom}
Kele Shao, Keda Tao, Can Qin, Haoxuan You, Yang Sui, and Huan Wang.
\newblock Holitom: Holistic token merging for fast video large language models.
\newblock In \emph{NeurIPS}, 2025.

\bibitem[Shao et~al.(2026)Shao, Tao, Zhang, Feng, Cai, Shang, You, Qin, Sui, and Wang]{shao2026tokens}
Kele Shao, Keda Tao, Kejia Zhang, Sicheng Feng, Mu Cai, Yuzhang Shang, Haoxuan You, Can Qin, Yang Sui, and Huan Wang.
\newblock A survey of token compression for efficient multimodal large language models.
\newblock \emph{TMLR}, 2026.

\bibitem[Shen et~al.(2025{\natexlab{a}})Shen, Gong, He, Zhang, Liu, Zhao, and Ding]{shen2025fastvid}
Leqi Shen, Guoqiang Gong, Tao He, Yifeng Zhang, Pengzhang Liu, Sicheng Zhao, and Guiguang Ding.
\newblock Fastvid: Dynamic density pruning for fast video large language models.
\newblock In \emph{NeurIPS}, 2025{\natexlab{a}}.

\bibitem[Shen et~al.(2025{\natexlab{b}})Shen, Xiong, Zhao, Wu, Chen, Zhu, Liu, Xiao, Varadarajan, Bordes, Liu, Xu, Kim, Soran, Krishnamoorthi, Elhoseiny, and Chandra]{shen2025longvu}
Xiaoqian Shen, Yunyang Xiong, Changsheng Zhao, Lemeng Wu, Jun Chen, Chenchen Zhu, Zechun Liu, Fanyi Xiao, Balakrishnan Varadarajan, Florian Bordes, Zhuang Liu, Hu Xu, Hyunwoo~J. Kim, Bilge Soran, Raghuraman Krishnamoorthi, Mohamed Elhoseiny, and Vikas Chandra.
\newblock Longvu: Spatiotemporal adaptive compression for long video-language understanding.
\newblock In \emph{ICML}, 2025{\natexlab{b}}.

\bibitem[Song et~al.(2026)Song, Chai, Ye, Hwang, Li, and Wang]{song2026moviechatplus}
Enxin Song, Wenhao Chai, Tian Ye, Jenq-Neng Hwang, Xi Li, and Gaoang Wang.
\newblock Moviechat+: Question-aware sparse memory for long video question answering.
\newblock \emph{IEEE TPAMI}, 48\penalty0 (1):\penalty0 374--389, 2026.

\bibitem[Tao et~al.(2025{\natexlab{a}})Tao, Qin, You, Sui, and Wang]{tao2025dycoke}
Keda Tao, Can Qin, Haoxuan You, Yang Sui, and Huan Wang.
\newblock Dycoke: Dynamic compression of tokens for fast video large language models.
\newblock In \emph{CVPR}, 2025{\natexlab{a}}.

\bibitem[Tao et~al.(2025{\natexlab{b}})Tao, Shao, Yu, Wang, Liu, and Wang]{tao2025omnizip}
Keda Tao, Kele Shao, Bohan Yu, Weiqiang Wang, Jian Liu, and Huan Wang.
\newblock Omnizip: Audio-guided dynamic token compression for fast omnimodal large language models.
\newblock \emph{arXiv preprint arXiv:2511.14582}, 2025{\natexlab{b}}.

\bibitem[Tao et~al.(2025{\natexlab{c}})Tao, You, Sui, Qin, and Wang]{tao2025plugandplay1xbitkvcache}
Keda Tao, Haoxuan You, Yang Sui, Can Qin, and Huan Wang.
\newblock Plug-and-play 1.x-bit kv cache quantization for video large language models.
\newblock \emph{arXiv preprint arXiv:2503.16257}, 2025{\natexlab{c}}.

\bibitem[Wan et~al.(2024)Wan, Wu, Liu, Huang, Zhu, Jin, Wang, and Yuan]{wan2024lookm}
Zhongwei Wan, Ziang Wu, Che Liu, Jinfa Huang, Zhihong Zhu, Peng Jin, Longyue Wang, and Li Yuan.
\newblock Look-m: Look-once optimization in kv cache for efficient multimodal long-context inference.
\newblock In \emph{Findings of EMNLP}, 2024.

\bibitem[Wang et~al.(2025{\natexlab{a}})Wang, Feng, Lai, Xu, Li, Ge, Dehghan, Cao, and Huang]{wang2025streambridge}
Haibo Wang, Bo Feng, Zhengfeng Lai, Mingze Xu, Shiyu Li, Weifeng Ge, Afshin Dehghan, Meng Cao, and Ping Huang.
\newblock Streambridge: Turning your offline video large language model into a proactive streaming assistant.
\newblock In \emph{NeurIPS}, 2025{\natexlab{a}}.

\bibitem[Wang et~al.(2024)Wang, Bai, Tan, Wang, Fan, Bai, Chen, Liu, Wang, Ge, Fan, Dang, Du, Ren, Men, Liu, Zhou, Zhou, and Lin]{wang2024qwen2vl}
Peng Wang, Shuai Bai, Sinan Tan, Shijie Wang, Zhihao Fan, Jinze Bai, Keqin Chen, Xuejing Liu, Jialin Wang, Wenbin Ge, Yang Fan, Kai Dang, Mengfei Du, Xuancheng Ren, Rui Men, Dayiheng Liu, Chang Zhou, Jingren Zhou, and Junyang Lin.
\newblock Qwen2-vl: Enhancing vision-language model's perception of the world at any resolution.
\newblock \emph{arXiv preprint arXiv:2409.12191}, 2024.

\bibitem[Wang et~al.(2025{\natexlab{b}})Wang, Gao, Gu, Pu, Cui, Wei, Liu, Jing, Ye, Shao, Wang, Chen, Zhang, Yang, Wang, Wei, Yin, Li, Cui, Chen, Ding, Tian, Wu, Xie, Li, Yang, Duan, Wang, Hou, Hao, Zhang, Li, Zhao, Duan, Deng, Fu, He, Wang, He, Shi, He, Xiong, Lv, Wu, Shao, Zhang, Deng, Qi, Ge, Guo, Zhang, Zhang, Cao, Lin, Tang, Gao, Huang, Gu, Lyu, Tang, Wang, Lv, Ouyang, Wang, Dou, Zhu, Lu, Lin, Dai, Su, Zhou, Chen, Qiao, Wang, and Luo]{wang2025internvl35}
Weiyun Wang, Zhangwei Gao, Lixin Gu, Hengjun Pu, Long Cui, Xingguang Wei, Zhaoyang Liu, Linglin Jing, Shenglong Ye, Jie Shao, Zhaokai Wang, Zhe Chen, Hongjie Zhang, Ganlin Yang, Haomin Wang, Qi Wei, Jinhui Yin, Wenhao Li, Erfei Cui, Guanzhou Chen, Zichen Ding, Changyao Tian, Zhenyu Wu, Jingjing Xie, Zehao Li, Bowen Yang, Yuchen Duan, Xuehui Wang, Zhi Hou, Haoran Hao, Tianyi Zhang, Songze Li, Xiangyu Zhao, Haodong Duan, Nianchen Deng, Bin Fu, Yinan He, Yi Wang, Conghui He, Botian Shi, Junjun He, Yingtong Xiong, Han Lv, Lijun Wu, Wenqi Shao, Kaipeng Zhang, Huipeng Deng, Biqing Qi, Jiaye Ge, Qipeng Guo, Wenwei Zhang, Songyang Zhang, Maosong Cao, Junyao Lin, Kexian Tang, Jianfei Gao, Haian Huang, Yuzhe Gu, Chengqi Lyu, Huanze Tang, Rui Wang, Haijun Lv, Wanli Ouyang, Limin Wang, Min Dou, Xizhou Zhu, Tong Lu, Dahua Lin, Jifeng Dai, Weijie Su, Bowen Zhou, Kai Chen, Yu Qiao, Wenhai Wang, and Gen Luo.
\newblock Internvl3.5: Advancing open-source multimodal models in versatility, reasoning, and efficiency.
\newblock \emph{arXiv preprint arXiv:2508.18265}, 2025{\natexlab{b}}.

\bibitem[Wang et~al.(2025{\natexlab{c}})Wang, Liu, Gui, Lin, Yang, Liao, Chen, and Zhang]{wang2025stc}
Yiyu Wang, Xuyang Liu, Xiyan Gui, Xinying Lin, Boxue Yang, Chenfei Liao, Tailai Chen, and Linfeng Zhang.
\newblock Accelerating streaming video large language models via hierarchical token compression.
\newblock \emph{arXiv preprint arXiv:2512.00891}, 2025{\natexlab{c}}.

\bibitem[Wang et~al.(2025{\natexlab{d}})Wang, Purushwalkam, Xiong, Savarese, Ji, and Xu]{wang2025dymu}
Zhenhailong Wang, Senthil Purushwalkam, Caiming Xiong, Silvio Savarese, Heng Ji, and Ran Xu.
\newblock Dymu: Dynamic merging and virtual unmerging for efficient variable-length vlms.
\newblock In \emph{NeurIPS}, 2025{\natexlab{d}}.

\bibitem[Wei et~al.(2025)Wei, Wan, Yu, Wang, Yang, Mao, Zhu, Cai, Wang, Chen, Liu, and Pang]{wei2025streamvln}
Meng Wei, Chenyang Wan, Xiqian Yu, Tai Wang, Yuqiang Yang, Xiaohan Mao, Chenming Zhu, Wenzhe Cai, Hanqing Wang, Yilun Chen, Xihui Liu, and Jiangmiao Pang.
\newblock Streamvln: Streaming vision-and-language navigation via slowfast context modeling.
\newblock \emph{arXiv preprint arXiv:2507.05240}, 2025.

\bibitem[Wu et~al.(2024)Wu, Li, Chen, and Li]{wu2024longvideobench}
Haoning Wu, Dongxu Li, Bei Chen, and Junnan Li.
\newblock Longvideobench: A benchmark for long-context interleaved video-language understanding.
\newblock In \emph{NeurIPS}, 2024.

\bibitem[Wu et~al.(2025{\natexlab{a}})Wu, Chen, Ming, Gao, Hu, He, and Yu]{wu2025totrlunlockllmtreeofthoughts}
Haoyuan Wu, Xueyi Chen, Rui Ming, Jilong Gao, Shoubo Hu, Zhuolun He, and Bei Yu.
\newblock Totrl: Unlock llm tree-of-thoughts reasoning potential through puzzles solving.
\newblock \emph{arXiv preprint arXiv:2505.12717}, 2025{\natexlab{a}}.

\bibitem[Wu et~al.(2025{\natexlab{b}})Wu, Ming, Gao, Zhao, Chen, Yang, Zheng, He, and Yu]{wu2025onpolicyoptimizationgroupequivalent}
Haoyuan Wu, Rui Ming, Jilong Gao, Hangyu Zhao, Xueyi Chen, Yikai Yang, Haisheng Zheng, Zhuolun He, and Bei Yu.
\newblock On-policy optimization with group equivalent preference for multi-programming language understanding.
\newblock In \emph{NeurIPS}, 2025{\natexlab{b}}.

\bibitem[Xing et~al.(2025)Xing, Huang, Dong, Lu, Zhang, Zang, Cao, He, Wang, Wu, and Lin]{xing2025pyramiddrop}
Long Xing, Qidong Huang, Xiaoyi Dong, Jiajie Lu, Pan Zhang, Yuhang Zang, Yuhang Cao, Conghui He, Jiaqi Wang, Feng Wu, and Dahua Lin.
\newblock Pyramiddrop: Accelerating your large vision-language models via pyramid visual redundancy reduction.
\newblock In \emph{CVPR}, 2025.

\bibitem[Xiong et~al.(2025)Xiong, Yang, Yu, Zhuge, Zhang, Zhu, and Lu]{xiong2025streamingvideounderstandingmultiround}
Haomiao Xiong, Zongxin Yang, Jiazuo Yu, Yunzhi Zhuge, Lu Zhang, Jiawen Zhu, and Huchuan Lu.
\newblock Streaming video understanding and multi-round interaction with memory-enhanced knowledge.
\newblock In \emph{ICLR}, 2025.

\bibitem[Xu et~al.(2026)Xu, Xiao, Chen, He, Peng, Lu, and Han]{xu2026streamingvlm}
Ruyi Xu, Guangxuan Xiao, Yukang Chen, Liuning He, Kelly Peng, Yao Lu, and Song Han.
\newblock Streamingvlm: Real-time understanding for infinite video streams.
\newblock In \emph{ICLR}, 2026.

\bibitem[Yang et~al.(2025{\natexlab{a}})Yang, Li, Yang, Zhang, Hui, Zheng, Yu, Gao, Huang, Lv, Zheng, Liu, Zhou, Huang, Hu, Ge, Wei, Lin, Tang, Yang, Tu, Zhang, Yang, Yang, Zhou, Zhou, Lin, Dang, Bao, Yang, Yu, Deng, Li, Xue, Li, Zhang, Wang, Zhu, Men, Gao, Liu, Luo, Li, Tang, Yin, Ren, Wang, Zhang, Ren, Fan, Su, Zhang, Zhang, Wan, Liu, Wang, Cui, Zhang, Zhou, and Qiu]{yang2025qwen3}
An Yang, Anfeng Li, Baosong Yang, Beichen Zhang, Binyuan Hui, Bo Zheng, Bowen Yu, Chang Gao, Chengen Huang, Chenxu Lv, Chujie Zheng, Dayiheng Liu, Fan Zhou, Fei Huang, Feng Hu, Hao Ge, Haoran Wei, Huan Lin, Jialong Tang, Jian Yang, Jianhong Tu, Jianwei Zhang, Jianxin Yang, Jiaxi Yang, Jing Zhou, Jingren Zhou, Junyang Lin, Kai Dang, Keqin Bao, Kexin Yang, Le Yu, Lianghao Deng, Mei Li, Mingfeng Xue, Mingze Li, Pei Zhang, Peng Wang, Qin Zhu, Rui Men, Ruize Gao, Shixuan Liu, Shuang Luo, Tianhao Li, Tianyi Tang, Wenbiao Yin, Xingzhang Ren, Xinyu Wang, Xinyu Zhang, Xuancheng Ren, Yang Fan, Yang Su, Yichang Zhang, Yinger Zhang, Yu Wan, Yuqiong Liu, Zekun Wang, Zeyu Cui, Zhenru Zhang, Zhipeng Zhou, and Zihan Qiu.
\newblock Qwen3 technical report.
\newblock \emph{arXiv preprint arXiv:2505.09388}, 2025{\natexlab{a}}.

\bibitem[Yang et~al.(2025{\natexlab{b}})Yang, Sui, Xiao, Huang, Gong, Li, Yan, Bai, Sadayappan, Hu, and Yuan]{yang2025topv}
Cheng Yang, Yang Sui, Jinqi Xiao, Lingyi Huang, Yu Gong, Chendi Li, Jinghua Yan, Yu Bai, Ponnuswamy Sadayappan, Xia Hu, and Bo Yuan.
\newblock Topv: Compatible token pruning with inference time optimization for fast and low-memory multimodal vision language model.
\newblock In \emph{CVPR}, 2025{\natexlab{b}}.

\bibitem[Yang et~al.(2025{\natexlab{c}})Yang, Chen, Tian, Wang, Li, Yu, and Jia]{yang2025visionzip}
Senqiao Yang, Yukang Chen, Zhuotao Tian, Chengyao Wang, Jingyao Li, Bei Yu, and Jiaya Jia.
\newblock Visionzip: Longer is better but not necessary in vision language models.
\newblock In \emph{CVPR}, 2025{\natexlab{c}}.

\bibitem[Yang et~al.(2025{\natexlab{d}})Yang, Zhao, Shukla, Singh, Mishra, Zhang, and Ren]{yang2025streammem}
Yanlai Yang, Zhuokai Zhao, Satya~Narayan Shukla, Aashu Singh, Shlok~Kumar Mishra, Lizhu Zhang, and Mengye Ren.
\newblock Streammem: Query-agnostic kv cache memory for streaming video understanding.
\newblock \emph{arXiv preprint arXiv:2508.15717}, 2025{\natexlab{d}}.

\bibitem[Yao et~al.(2025)Yao, Li, Wei, Li, Ren, Liu, Ouyang, Wang, Li, Li, Kong, Liu, Zhang, and Sun]{yao2025timechat}
Linli Yao, Yicheng Li, Yuancheng Wei, Lei Li, Shuhuai Ren, Yuanxin Liu, Kun Ouyang, Lean Wang, Shicheng Li, Sida Li, Lingpeng Kong, Qi Liu, Yuanxing Zhang, and Xu Sun.
\newblock Timechat-online: 80\% visual tokens are naturally redundant in streaming videos.
\newblock In \emph{ACM MM}, 2025.

\bibitem[Zhai et~al.(2023)Zhai, Mustafa, Kolesnikov, and Beyer]{zhai2023siglip}
Xiaohua Zhai, Basil Mustafa, Alexander Kolesnikov, and Lucas Beyer.
\newblock Sigmoid loss for language image pre-training.
\newblock In \emph{ICCV}, 2023.

\bibitem[Zhang et~al.(2025{\natexlab{a}})Zhang, Wang, Tang, Liu, Feng, and Jin]{zhang2025flash}
Haoji Zhang, Yiqin Wang, Yansong Tang, Yong Liu, Jiashi Feng, and Xiaojie Jin.
\newblock Flash-vstream: Efficient real-time understanding for long video streams.
\newblock In \emph{ICCV}, 2025{\natexlab{a}}.

\bibitem[Zhang et~al.(2025{\natexlab{b}})Zhang, Li, Zhang, Pu, Cahyono, Hu, Liu, Zhang, Yang, Li, and Liu]{zhang2025lmmseval}
Kaichen Zhang, Bo Li, Peiyuan Zhang, Fanyi Pu, Joshua~Adrian Cahyono, Kairui Hu, Shuai Liu, Yuanhan Zhang, Jingkang Yang, Chunyuan Li, and Ziwei Liu.
\newblock Lmms-eval: Reality check on the evaluation of large multimodal models.
\newblock In \emph{Findings of NAACL}, 2025{\natexlab{b}}.

\bibitem[Zhang et~al.(2025{\natexlab{c}})Zhang, Wu, Li, Li, Ma, Liu, and Li]{zhang2025llavavideo}
Yuanhan Zhang, Jinming Wu, Wei Li, Bo Li, Zejun Ma, Ziwei Liu, and Chunyuan Li.
\newblock Llava-video: Video instruction tuning with synthetic data.
\newblock \emph{TMLR}, 2025{\natexlab{c}}.

\bibitem[Zhou et~al.(2025{\natexlab{a}})Zhou, Shu, Zhao, Wu, Liang, Xiao, Qin, Yang, Xiong, Zhang, Huang, and Liu]{zhou2025mlvu}
Junjie Zhou, Yan Shu, Bo Zhao, Boya Wu, Zhengyang Liang, Shitao Xiao, Minghao Qin, Xi Yang, Yongping Xiong, Bo Zhang, Tiejun Huang, and Zheng Liu.
\newblock Mlvu: Benchmarking multi-task long video understanding.
\newblock In \emph{CVPR}, 2025{\natexlab{a}}.

\bibitem[Zhou et~al.(2025{\natexlab{b}})Zhou, Ruan, Ling, Chen, Wang, and Jiang]{zhou2025tvc}
Lebin Zhou, Cihan Ruan, Nam Ling, Zhenghao Chen, Wei Wang, and Wei Jiang.
\newblock Tvc: Tokenized video compression with ultra-low bit rate.
\newblock \emph{Visual Intelligence}, 3:\penalty0 25, 2025{\natexlab{b}}.

\bibitem[Zhu et~al.(2026)Zhu, Wang, Su, Wang, and Wang]{zhu2026obsdiff}
Junhan Zhu, Hesong Wang, Mingluo Su, Zefang Wang, and Huan Wang.
\newblock Obs-diff: Accurate pruning for diffusion models in one-shot.
\newblock In \emph{ICLR}, 2026.

\end{thebibliography}
}

\clearpage
\setcounter{page}{1}
\maketitlesupplementary
\appendix

\section{Implementation Details}

\subsection{Terminology Clarification}

Throughout this paper, \textit{Pre-LLM} and \textit{Post-LLM} refer to \textbf{pipeline location}, not inference stage.
\textbf{Pre-LLM} denotes compression applied after the vision encoder and projector but before visual tokens enter the LLM transformer layers.
\textbf{Post-LLM} denotes memory management applied after kv-cache formation within the LLM.
Both operations occur during the prefill phase; the distinction captures \textit{where} in the pipeline compression acts, not \textit{when} during inference.

\subsection{Streaming Attention and Token Saliency}

CTR necessitates a saliency score for each visual token derived from the vision encoder.
Rather than materializing full attention maps, we implement a streaming self-attention kernel inspired by memory-efficient chunked attention~\citep{dao2024flashattention2}, integrated into the SigLIP encoder~\citep{zhai2023siglip} to compute both the attention output and per-token importance scores simultaneously.
Given queries $Q$, keys $K$, values $V$, and an optional mask $M$, the kernel processes keys in small chunks through two passes over the sequence.
The first pass accumulates a numerically stable log-sum-exp of the attention scores along the key dimension, while the second pass recomputes the scores to obtain normalized probabilities.
These probabilities are subsequently aggregated into the final output and an importance accumulator.
Algorithm~\ref{alg:streaming-attn-importance} summarizes this procedure.

\begin{algorithm}[t]
\caption{Streaming attention with importance scores}
\label{alg:streaming-attn-importance}
\begin{algorithmic}[1]
\REQUIRE Query $Q$, keys $K$, values $V$, optional mask $M$, scale $\alpha$, chunk size $C$
\ENSURE Attention output $O$, importance scores $s$
\STATE Convert $Q,K,V$ to \texttt{float32} and initialize $\text{attn\_max}$, $\text{attn\_lse}$, $O$, $\text{imp}$
\FOR{each key chunk $[s:e)$ of size at most $C$}
    \STATE Compute scores $S = \alpha Q K_{s:e}^\top$
    \IF{$M$ is provided}
        \STATE Add corresponding mask slice: $S \gets S + M_{:, :, :, s:e}$
    \ENDIF
    \STATE Update $\text{attn\_max}$ and $\text{attn\_lse}$ using log-sum-exp over the key dimension
\ENDFOR
\FOR{each key chunk $[s:e)$}
    \STATE Recompute $S = \alpha Q K_{s:e}^\top$
    \IF{$M$ is provided}
        \STATE Add corresponding mask slice: $S \gets S + M_{:, :, :, s:e}$
    \ENDIF
    \STATE Compute probabilities $P$ from $\text{attn\_max}$ and $\text{attn\_lse}$
    \STATE $O \gets O + P V_{s:e}$
    \STATE $\text{imp}_{s:e} \gets \text{imp}_{s:e} + \sum_{h,q} P$
\ENDFOR
\STATE $s \gets \text{imp} / (H \cdot L_q)$
\RETURN $O, s$
\end{algorithmic}
\end{algorithm}

Here $H$ denotes the number of attention heads and $L_q$ the number of query positions in the self-attention module.
We enable this streaming kernel exclusively on the final layer of the vision encoder to expose the resulting importance scores as an additional one-dimensional tensor over visual tokens.
CTR consumes these scores as attention-based saliency signals to rank, select, and compress visual tokens in a strictly streaming fashion without revisiting past frames.

\subsection{Text--Video Retrieval}

\paragraph{Layer-wise vision memory.}
During video encoding, we extract key tensors $K^{(l)} \in \mathbb{R}^{H_k \times T^{(l)} \times d}$ from every transformer block prior to applying rotary embeddings to ensure they contain exclusively visual tokens.
For each contiguous group of $G$ visual tokens, we compute a summary key
\[
k_{t}^{(l)} = \frac{1}{G} \sum_{i \in \mathcal{G}_{t}} K_{i}^{(l)}, \qquad t = 1,\dots,\frac{T^{(l)}}{G},
\]
and store the resulting sequence $\{k_t^{(l)}\}$ alongside the exact KV cache for that layer.
This process yields a layer-aligned memory bank that preserves the original attention-head dimensionality while maintaining the chronological order of visual segments.

\paragraph{Question-guided single-pass retrieval.}
Upon receiving a question, we exclude vision placeholders and prompts to process the clean text through the language model in a single pass.
At transformer layer $l$, we map the resulting query tensor $Q^{(l)} \in \mathbb{R}^{H_q \times T_q \times d}$ to the key--value head space by averaging query heads within each GQA group.
Subsequently, we average across the token dimension to obtain a layer-specific question embedding
\[
q^{(l)} = \frac{1}{T_q} \sum_{j=1}^{T_q} Q^{(l)}_{:,j,:} \in \mathbb{R}^{H_k \times d}.
\]
Before retrieval, we flatten both $q^{(l)}$ and each $k_t^{(l)}$ into vectors, $\ell_2$-normalize them, and perform a cosine-similarity top-$K$ search within that layer.
We determine the number of retrieved segments via
\[
K = \min\!\Bigl(\Bigl\lceil \frac{B}{G} \Bigr\rceil,\; \frac{T^{(l)}}{G}\Bigr),
\]
where $B$ represents the global token budget and $T^{(l)}$ denotes the total number of visual tokens stored at layer $l$.
We maintain the selected indices in sorted order to ensure the temporal coherence of the retrieved segments.

\paragraph{Cache reconstruction for decoding.}
Following this single forward pass, each layer identifies the necessary visual groups.
We retrieve the corresponding KV slices and system prompt tokens from the memory bank to assemble a \texttt{DynamicCache}, reusing these per-layer indices throughout the decoding process.
Consequently, the system incurs the question--video matching cost only once per question while permitting each layer to independently select the video segments contributing to its attention computation.

\subsection{RVS Streaming Evaluation}

While most benchmarks employ a conventional offline evaluation via the lmms-eval framework~\citep{zhang2025lmmseval} where one video--question pair is processed at a time with full clip access, RVS operates under a strict streaming protocol.
We sort all questions for a given video by their end timestamps.
With frames sampled at a fixed rate such as $0.5$\,fps, the model processes the video linearly and observes the content only once.
Given a question ending at time $t$, we encode only the new frames arriving between the last encoding step and time $t$ to update the streaming KV cache without re-encoding earlier frames.
Should a subsequent question fall within the timeframe of the previously encoded frames, we answer it using the existing cache and pass an empty image batch to indicate no additional visual input.
This protocol ensures that the model generates predictions solely from video content observed up to the question time, highlighting the streaming nature of our method on RVS.

\section{Additional Results}

\subsection{LongVideoBench}

\begin{table}[h]
\centering
\caption{Results on LongVideoBench.}\label{tab:longvideobench}
\begin{tabular}{llc}
\toprule
\textbf{Method} & \textbf{Setting} & \textbf{Accuracy} \\
\midrule
LLaVA-OV-7B & 32 frames & 56.4 \\
+StreamingTOM (ours) & 0.5 FPS & 56.3 \\
\bottomrule
\end{tabular}
\end{table}

We evaluate StreamingTOM on LongVideoBench~\citep{wu2024longvideobench}, which spans a wide duration range (8s--60min).
StreamingTOM retains 99.8\% of the baseline accuracy (56.3 vs.\ 56.4).
The marginal gap is expected, as LongVideoBench's duration distribution skews shorter than VideoMME-Long and MLVU, where StreamingTOM's streaming temporal coverage provides the greatest advantage (Table~\ref{tab:main_eval}).

\subsection{Generalization to Other VLMs}

To validate backbone-agnostic generalization, we apply StreamingTOM to Qwen2.5-VL-7B-Instruct~\citep{bai2025qwen25vl} on VideoMME.
The core CTR and OQM mechanisms require no modification; only minor adjustments to the input format and 3D rotary position encoding are needed.

\begin{table}[h]
\centering
\caption{Generalization results on Qwen2.5-VL-7B-Instruct.}\label{tab:qwen_generalization}
\resizebox{\linewidth}{!}{
\begin{tabular}{ll|cccc|c}
\toprule
\multirow{2}{*}{\textbf{Method}} & \multirow{2}{*}{\textbf{Setting}} & \multicolumn{4}{c|}{\textbf{VideoMME}} & \multirow{2}{*}{\textbf{Retain}} \\
& & Short & Medium & Long & Avg & \\
\midrule
Qwen2.5-VL-7B & 32 frames & 72.4 & 61.3 & 50.3 & 61.3 & 100\% \\
+StreamingTOM & 32 frames & 69.3 & 57.2 & 50.1 & 58.9 & 96.1\% \\
+StreamingTOM & 0.5 FPS & 69.9 & 65.2 & 54.0 & 63.0 & 102.8\% \\
\bottomrule
\end{tabular}
}
\end{table}

At 32 frames, StreamingTOM retains 96.1\% of baseline accuracy with substantially fewer tokens.
Under streaming settings (0.5 FPS), temporal coverage boosts retention to 102.8\%, consistent with our LLaVA-OV-7B findings (Table~\ref{tab:main_eval}), confirming backbone-agnostic generalization without modifying the core pipeline.

\section{Discussion}

\subsection{Adaptivity of StreamingTOM}

StreamingTOM capitalizes on the observation that long video sequences possess strong temporal locality and redundancy where nearby frames share content and few tokens drive downstream reasoning.
CTR and OQM harness this structure within a strictly training-free paradigm.
Since both modules operate exclusively on token representations and KV caches without assuming specific architectural details beyond a patch-based vision encoder and a transformer-style language model, StreamingTOM seamlessly integrates with diverse video LLMs.
This compatibility applies to both offline and streaming settings and requires no modification to the original training procedure.
The configurable per-frame token budget and global KV-cache budget serve as precise control mechanisms to balance accuracy, latency, and memory usage across varied deployment scenarios.

\subsection{Deployment Scenarios}

StreamingTOM's bounded memory growth and predictable latency make it applicable to several practical streaming scenarios.
\textbf{(1) Long-duration monitoring.}
In applications such as surveillance and meeting assistants, the kv-cache remains bounded regardless of stream length (e.g., 1.2\,GB for one hour of video), enabling continuous operation on a single GPU without memory overflow.
\textbf{(2) Latency-sensitive interaction.}
For embodied AI and autonomous driving assistants, sub-0.4s TTFT (Table~\ref{tab:efficiency_centered} in the main paper) supports real-time querying over live video streams.
\textbf{(3) Resource-constrained deployment.}
The fixed per-frame budget $G$ ensures predictable compute and memory consumption per frame, facilitating capacity planning on edge devices with limited resources.
Being training-free and backbone-agnostic, StreamingTOM can be integrated into existing VLM pipelines without retraining, lowering the barrier for practical adoption across diverse deployment environments.

\section{Limitations and Future Work}

Despite the compression and efficiency gains achieved in a training-free setting, the underlying multimodal backbone fundamentally constrains the overall performance of StreamingTOM.
Compression or memory management algorithms alone cannot resolve intrinsic limitations regarding fine-grained perception or long-range reasoning.
Adopting stronger or task-adapted backbones would directly raise the performance ceiling, making the joint tuning of the backbone with CTR and OQM a logical extension.
Furthermore, while our design optimizes prefill and bounds the active KV cache, the vision encoder remains the dominant latency source.
Consequently, dense feature extraction in the visual front-end limits end-to-end acceleration even when the language model executes with high efficiency.

Several avenues exist to extend StreamingTOM and address these limitations.
One path moves beyond purely training-free compression to explore lightweight tuning of CTR and OQM alongside the backbone by learning content-aware token budgets or memory allocation policies atop a frozen or partially fine-tuned video LLM.
Simultaneously, reducing the computational cost of the vision encoder warrants investigation.
This could involve introducing lightweight pre-encoder modules for early downsampling and motion-aware frame selection or integrating encoder-side sparsity and quantization techniques.
Integrating such backbone improvements with StreamingTOM offers the potential to further minimize the end-to-end cost of long-form video understanding while preserving robust performance.

\end{document}